\documentclass{article}

\usepackage[panel]{rival}

\usepackage[utf8]{inputenc}
\usepackage[T1]{fontenc}
\usepackage{url}
\usepackage{booktabs}
\usepackage{graphicx}
\usepackage{amsmath}
\usepackage{amssymb}
\usepackage[hidelinks]{hyperref}
\hypersetup{pdftitle={Taxi: Analyzing Agents at (Real) Scale}, pdfauthor={Guy Dar}}
\usepackage{xcolor}
\usepackage{listings}
\definecolor{yamlkey}{RGB}{0,80,160}
\definecolor{yamlcomment}{RGB}{100,130,100}
\definecolor{yamlstring}{RGB}{160,60,20}
\lstdefinelanguage{yaml}{
  keywords={true,false,null},
  keywordstyle=\color{yamlstring}\bfseries,
  sensitive=false,
  basicstyle=\color{yamlkey}\small\ttfamily,
  identifierstyle=\color{yamlkey},
  comment=[l]{\#},
  commentstyle=\color{yamlcomment}\itshape,
  stringstyle=\color{yamlstring},
  morestring=[b]',
  morestring=[b]",
}
\makeatletter
\newcommand{\taxi@toctype}{toc}
\newcommand{\taxi@dualcontentsline}[3]{%
  \taxi@addcontentsline{#1}{#2}{#3}%
  \def\taxi@thistype{#1}%
  \ifx\taxi@thistype\taxi@toctype
    \taxi@addcontentsline{apc}{#2}{#3}%
  \fi}
\newcommand{\appendixcontents}{%
  \section*{Appendix Contents}%
  \let\taxi@addcontentsline\addcontentsline
  \let\addcontentsline\taxi@dualcontentsline
  \@starttoc{apc}%
}
\makeatother

\title{Taxi: Analyzing Agents at (Real) Scale\\
\large Technical Report}

\author{%
  Guy Dar \\
  Rival Research \\
  \texttt{guydar@rivalsecurity.io}
}

\begin{document}

\maketitle

\begin{abstract}
In this technical report, we present \emph{Taxi} -- a system for debugging, clustering, and monitoring agents at scale. Taxi scales easily to millions of steps efficiently, without spurious expenses. 
This is made possible by a combination of scalable analysis methods. Taxi relies primarily on clustering to generate a taxonomy over steps for a structured overview of the data, and provides full SQL access for engineers and automated agents to ask complex questions. Clustering and querying are best used together in a hybrid pipeline guided by a coding agent.
Taxi also includes an experimental integration of Jev-style decision models: models that map state and bounded questions directly to typed, probabilistic decisions. They add low-cost semantic routing and evaluation where embeddings or literal queries alone are insufficient.
In addition, Taxi provides an extensible ecosystem, including built-in evaluators and daily reports.  
To accommodate growth, Taxi is designed to be modular, hackable, and easy to plug-and-play for individuals and teams.
\end{abstract}

\section{Introduction}
\label{sec:introduction}

Agents now run for hours or days \citep{kwa2026measuring}, chaining together planning,
tool use, and error recovery across long trajectories. Every run leaves a rich record of what the agent did that goes beyond just success and failure. Agent runs contain a costly yet elusive class of problems. To name a few examples: behavioral bugs that never raise an error, agents that quietly exceed the scope of what they were asked to do, the same unnecessary step repeated across thousands of
independent runs, unexpected tool results, and other undesired phenomena.  These phenomena are not necessarily easy to predict in advance. Some failure modes lead the agent astray, resulting in wrong conclusions \citep{zhu2025where}. Others may be local and recoverable, but they still come at the cost of extended work \citep{hu2026analyzing}. At production scale, a single
undetected recurring ``soft failure'' pattern is a burden, whether it is an actual mistake or just wasted tokens \citep{cuadron2025danger, bai2026how}. The solution to both lies in observability. 

Trajectory logging is common practice today. But for real observability, we contend there are, at minimum, two more things we need: (a) \textit{structured overview} of agent behavior across the corpus, and (b) letting users \textit{query} that behavior and ask arbitrarily complex questions. Modern products are emerging that fulfill these criteria to a certain degree \citep{laminar2026, meng2025docent, tannyhill2026langsmith}. Unfortunately, they all rely heavily on language models touching every trajectory (or a limited sample, for affordability), which is prohibitive at production volumes, due to costs, rate limits, latency, and limited coverage. Research pipelines that induce taxonomies with LLMs similarly have a model read each trajectory or a sample of them \citep{tamkin2024clio, cemri2026adaptive, gupta2025unsupervised}.

Taxi is built with scalability as a first-class citizen. Even with a production-grade workload and relatively limited resources, Taxi won't falter. In fact, the methods we use encourage granular visibility into agent trajectories, giving rise to a better perspective on the agents' work. Its two established analysis paths are cheap clustering and direct text querying. Taxi now adds an experimental third path based on Jev-style decision models (Section~\ref{sec:design}). These models accept a state and bounded questions and return typed decisions with probabilities rather than generated text \citep{almeida2026systemone}. In this report, \emph{Jev-style} refers to this broader interface, not only TypeSafe AI's original Jev model. Its substantially lower per-decision cost makes semantic judgment applicable at a different scale than general-purpose generation. For structured overview, Taxi builds a \textit{taxonomy}, a tree of hierarchical clusters, over the steps in the data. Then, each step is assigned to one of the clusters. Building on that, a user or agent can track information about the trajectories and get a better view of the data before analyzing it. In the following, we present the main aspects of Taxi's design. For the interested reader, we provide full details in the Appendix. Appendix~\ref{app:sastbench} walks through Taxi on one public agent benchmark, layer by layer.
Taxi is designed for individuals and teams, with out-of-the-box support for a range of sophisticated features, detailed below. Upon community interest, we will open source it.

\begin{figure}[t]
\centering
\includegraphics[width=0.88\linewidth]{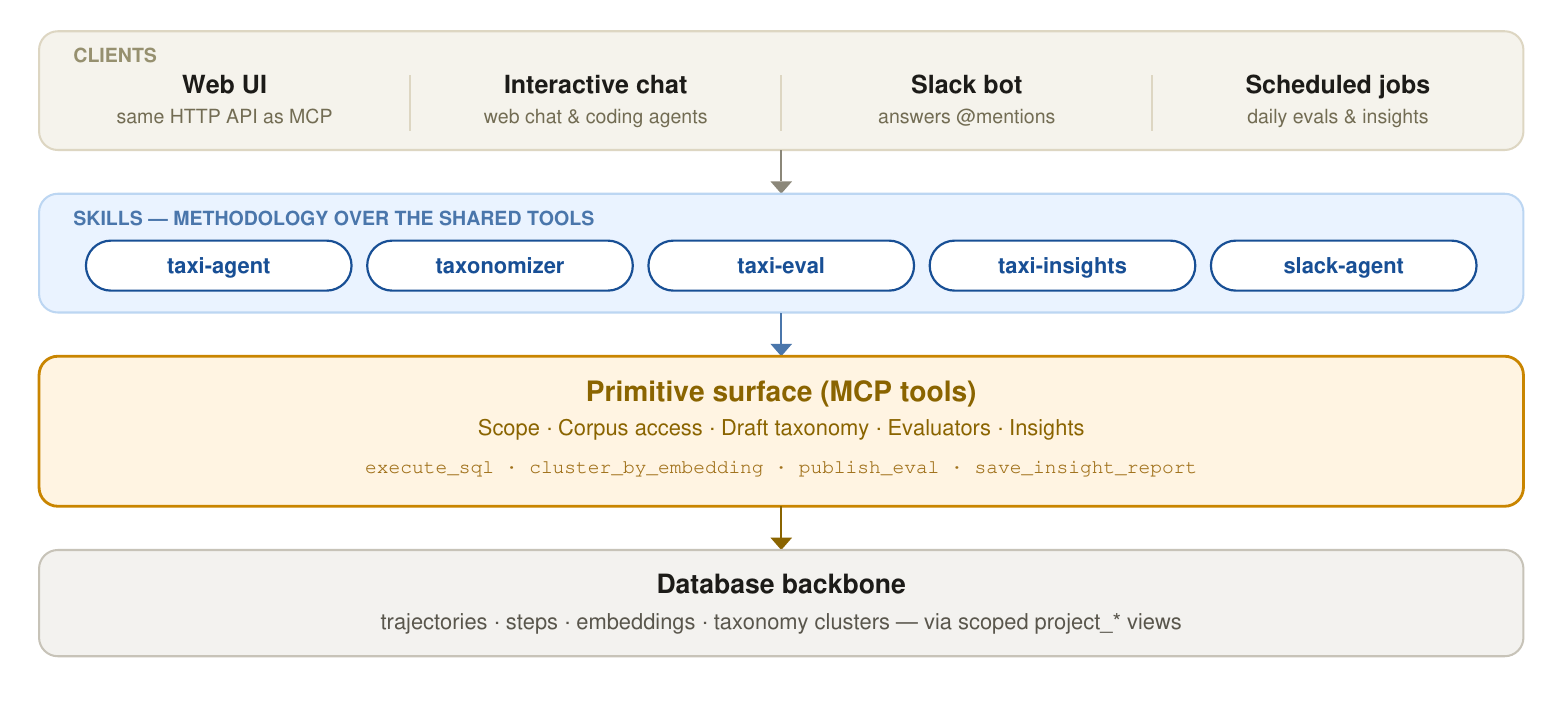}
\caption{Taxi architecture: clients share one primitive surface (MCP tools) over one database
backbone, with skills as the methodology layer that steers agents over those tools.}
\label{fig:architecture}
\end{figure}
\section{The Taxi System}

\subsection{Data Layer}
\label{sec:backbone}
\textbf{Data sources.} A \textbf{data source} holds a team's corpus of \emph{trajectories}
(complete agent runs). Each trajectory is an ordered sequence of \emph{steps} -- messages, tool
calls, and observations. A text embedding is computed and assigned to each step. The corpus is loaded from an
external origin through a \emph{loader} (LangSmith, Hugging Face, JSON export, and similar). All
loaders write the same schema, so supporting a new origin just means adding a loader. Loader-specific extra information is captured under the metadata field. Data sources are shared across the team -- everyone reads the same trajectories
and embeddings, and nobody re-ingests the same logs for their own copy. Origins that keep producing new runs (LangSmith is one) are \emph{syncable}: each sync bumps a monotonic data version, and every trajectory records which version first contained it.

\textbf{Projects.} A \textbf{project} is a personal view over one data source. It belongs to a
single user and can apply its own filters (e.g., a project filtered by a specific run name). That split lets the team pay the cost of ingestion once, while each person keeps a separate workspace. The main artifact in a project is a \textbf{taxonomy}: a hierarchical tree of step clusters (Section~\ref{sec:taxonomizer}).
Each cluster has a short label and description, and each step is assigned to one cluster, or left unassigned when no rule matches. Taxonomies are versioned and immutable once published -- editing the tree means publishing a new taxonomy version. After a data-source sync, a project can pull the new trajectories and assign their steps under its current taxonomy version; the cluster definitions and assignment rules stay as they were, and only membership of new steps is filled in. Decoupling data evolution (via data versions) and taxonomy evolution is desired to keep the two orthogonal -- the steps of an old trajectory can be discretized into new clusters, and a new trajectory can be discretized with an old taxonomy.

\subsection{Design}
\label{sec:design}

\textbf{Architecture.} Taxi is organized in three layers (Figure~\ref{fig:architecture}). At the bottom there's a shared
database of trajectories broken into steps, with embeddings and (when present) a taxonomy of step clusters. All the other layers use the database. 
The middle layer is a set of a few dozen
\textbf{tools} exposed over MCP \citep{anthropic2024modelcontextprotocol}. These tools support querying a project, building or editing a taxonomy draft, publishing evaluators and insight definitions. 
On top sit \textbf{skills} \citep{anthropic2025agentskills} -- short instruction documents
that tell a coding agent or chat model which tools to use, and how, for a particular job (explore
the data, build a taxonomy, write an eval, produce a daily report). 

All clients, web UI, scheduled jobs, and others, 
use the same pool of tools. The point of the split is that analysis methods change faster than the data model -- a new way to taxonomize or a new kind of report is usually a new skill or a tool, not a rewrite of storage or every client.

\paragraph{Scaling.} Taxi combines two established methods with a Jev-style path:
\begin{itemize}
    \item \textbf{Embedding and Clustering} -- vector clustering over embeddings is cheap and scales easily enough to run over every step in the corpus \citep{arya1993ann,douze2024faiss}, giving us a structured overview with no LLM calls (Section~\ref{sec:taxonomizer}).
    \item \textbf{Querying and Textual Regularity} -- much of agent behavior is searchable as plain text: tool names, error strings, argument shapes. SQL and pattern matching over steps answer a surprisingly wide range of questions  with no need for AI (Section~\ref{sec:sql}). 
    \item \textbf{Jev-style Decision Models} -- when a category is semantic rather than geometric or lexical, a decision model can answer a bounded choice or yes/no question directly, returning probabilities without generating an explanation. Taxi uses this interface both to route steps through taxonomy levels and to evaluate trajectories. Its cost profile permits semantic decisions to be applied far more broadly than generative-model calls, giving it the potential to become the main approach for semantic analysis.
\end{itemize}
The first two methods remain Taxi's established core. When a remaining task requires a generative-model judgment, Taxi can use clusters and queries to narrow the relevant set before sampling it; this is a cost-control tactic rather than a separate analysis approach. Table~\ref{tab:routing} compares the three routing methods, with a generative LLM for reference; all scale linearly in the number of steps.

\begin{table}[t]
\centering
\small
\begin{tabular}{@{}lcccc@{}}
\toprule
 & Cost per step & After a taxonomy edit & Coverage & Category defined by \\
\midrule
Embedding & low, once & cheap & full & proximity \\
Regex & none & cheap & partial & surface form \\
Jev-style & medium & rerun model & full & definition \\
Generative LLM & high & rerun model & full & definition \\
\bottomrule
\end{tabular}
\caption{Routing steps into a taxonomy.}
\label{tab:routing}
\end{table}

\subsection{Taxonomy Construction}
\label{sec:taxonomizer}
Taxi's design supports switching and composing assignment methods. We present a flat clustering pipeline and an agentic construction pipeline; the latter can mix three routing mechanisms within one tree.
\paragraph{Simple Flat Clustering.} That's the simplest form of taxonomy construction. We run a clustering algorithm, e.g. k-means \citep{lloyd1982least}, on the embeddings. The resulting clusters then form the taxonomy. Naming the clusters takes place by calling an LLM on a small sample of steps from each cluster. Upon sync, new steps are simply assigned to the cluster which they are closest to. Despite its simplicity, these clusters tend to be useful for understanding and debugging agents. 

\paragraph{Agentic Taxonomy Construction.} This method can be used by any agentic harness. It builds and refines the cluster tree, rather than running a single clustering pass. The procedure is an \textit{agent skill}: a short instruction document that steers the model over relevant MCP tools. The full details of the pipeline are provided in Appendix~\ref{app:taxonomizer}. The main pieces are:

\begin{itemize}
  \item \textbf{Corpus exploration} -- SQL over the scoped views, plus TF-IDF term lists that
    surface recurring vocabulary in the steps (and, when useful, inside clusters).
  \item \textbf{Embedding-based clustering} -- k-means over step embeddings to propose a partition
    at the root or under an existing cluster.
  \item \textbf{Regex-based clustering} -- clusters defined by patterns over step text, when a
    behavior has a stable surface form; We defer more details to Appendix~\ref{app:taxonomizer}.
  \item \textbf{Jev-based routing} -- clusters defined by natural-language labels and
    descriptions. A decision model chooses among the sibling definitions, with an explicit
    ``none of these'' option and a confidence threshold. This is useful when membership is
    semantic but the categories are already known.
  \item \textbf{Jev-based construction} (experimental) -- discovers Jev categories from the
    corpus instead of taking them as given: the decision model makes every decision and an LLM
    only writes the categories it decides to change (Appendix~\ref{app:jev-construction}).
  \item \textbf{Scoped refinement} -- the above tools can be restricted to one cluster's members
    (TF-IDF inside a cluster, split by another k-means call, merge near-duplicates, and so on), so
    the agent can iterate until clusters are interpretable and distinct.
\end{itemize}

\textbf{Discretization.} Apart from being a goal unto itself, taxonomization plays a few more roles. First, it provides a starting point for an agent or human analyzing the dataset, giving them context for understanding the domain and behaviors worth looking into during query-based work (Section~\ref{sec:sql}). Second, it \textit{discretizes} the steps -- breaks them into a finite set of discrete clusters. This opens up a new world of possibilities -- now trajectories can be interpreted as coming from a \textit{state machine}, and the transition map can be analyzed, and tools for analyzing discrete trajectories can be used, including tools from process mining \citep{van2011process, van2004process}, motif analysis \citep{bailey1994fitting}, and anomaly detection \citep{chandola2009anomaly}.



\begin{figure}[t]
\centering
\includegraphics[width=0.88\linewidth]{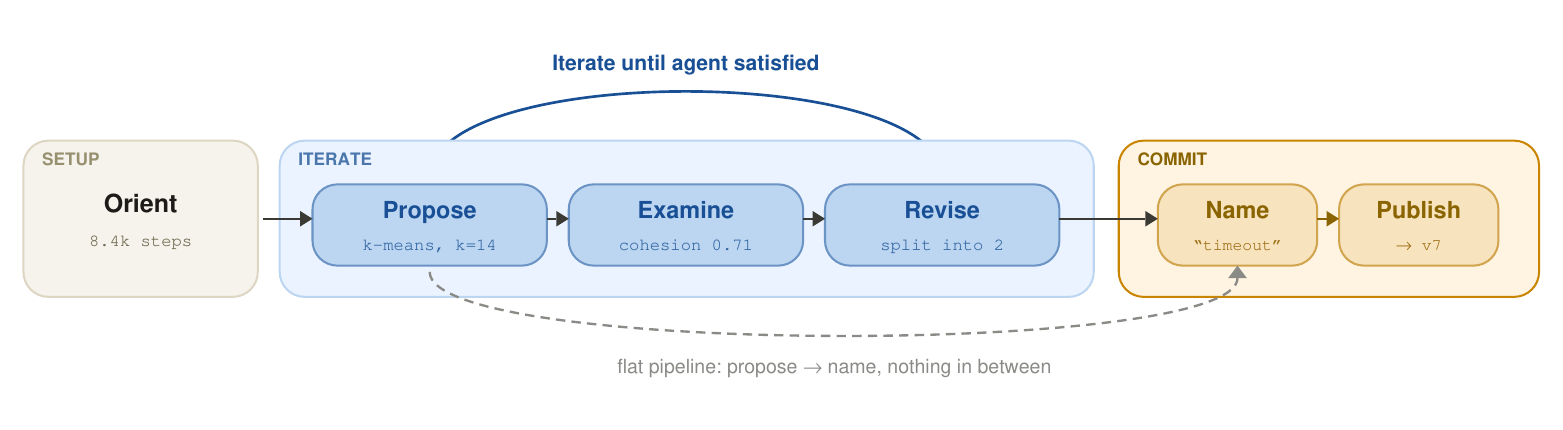}
\caption{Taxonomy construction as the taxonomizer skill runs it. The clustering call proposes, the
examination step decides. The dashed shortcut is the \textit{flat} pipeline, which skips examination.}
\label{fig:taxonomy-construction}
\end{figure}

\subsection{Querying the Corpus and Textual Regularity}
\label{sec:sql}

A taxonomy gives a birdseye view. It does not replace asking arbitrary questions about what agents
did. Our trust in the power of text-based exploration is not speculative. Modern coding agents rely on exact text and regex-based tools to explore code all the time \citep{yang2024sweagent}, and recent work has shown that \textit{grep} is often stronger than embeddings \citep{sen2026grep}. In a way, this indicates that text is more structured than one might assume, and the same phrases are used for similar things. In Taxi, we nickname that the ``regularity of text''. Taxi keeps the full text of every step, so a human or a tool-using LLM can search and filter the
corpus directly -- e.g., how often a tool error appears, which runs hit a given cluster,
whether a phrase shows up in final answers, or how behavior differs across models or time.

SQL queries don't usually need access to the raw tables in the database. Before executing any query, a \textbf{scope} is defined (also called a  \textit{view}): either
a project (a data source with that project's filters and latest taxonomy) or one or more data
sources (raw corpus, no taxonomy). Queries then run as read-only SQL inside this scope, given easy access to the scope's trajectories, steps, and, under a project, also clusters
and step-to-cluster assignments. This all happens automatically when an engineer or AI agent uses the \verb|execute_sql| tool (scope must be set first by using \verb|set_project|/\verb|set_data_sources|). Agents, evaluators, and insight reports all use the same path.

\subsection{Ecosystem}
\label{sec:ecosystem}

Until now, we have discussed the core of Taxi, which is based on a unified trajectory-step-embedding database and clustering procedures. On top of that common ground, many extensions can be designed. Below we describe two such built-in extensions that we designed. Due to the ease of developing such extensions with little concern to the underlying core, we view them as Taxi's \textbf{ecosystem}. 

\textbf{Evaluators} (Appendix~\ref{app:eval-yaml}) are regression checks over trajectories. Once a
behavioral bug is understood -- an agent that skips a required tool, leaks a secret into a
user-facing report, or repeats a pattern that was already fixed -- the team writes a check that
fails when that pattern shows up again. This way, past bug fixes can be turned into checks. A coding agent can even create the evaluation check on its own. If given access to recent code changes, it can identify bug fixes in the code and design an evaluator automatically. Evaluator criteria may be SQL or regex queries, or a Jev-style yes/no question applied per trajectory or per step. SQL remains useful for cheap narrowing before semantic judgment.

\textbf{Insight reports} (Appendix~\ref{app:insights}) answer a different question. It focuses on periodically keeping an eye on recently generated trajectories. It finds surprising or otherwise relevant agent behavior in trajectories created after the last report. This complements other approaches that focus on the aggregate behavior. Insights can run as a scheduled routine. Findings focus around how the agents acted, but in some cases can highlight interesting outcomes of the inspected agents themselves.

\section{Limitations and Outlook}
\label{sec:discussion} 
Taxi relies on the correctness of its clusters. Clustering long text is hard -- steps may contain long code blocks that do not contribute to the axes we care about, and hyperparameters are hard to set well. The agentic approach improves precision substantially but can still miss nuances. Trajectory behavior may also drift over time \citep{gama2014survey}; an old taxonomy can force a stale lens on new data. Jev-style routing and evaluator criteria have only recently been integrated into Taxi and have not yet been tested end to end across the full system; for now we therefore mark these paths experimental. Finally, Taxi does not yet handle multi-agent systems out-of-the-box. 
These are limitations of Taxi's \textit{current behavior}, not its architecture. Its modularity can accommodate multi-agent support, better taxonomization instructions, and other extensions without redesign. Taxi is fully operational and under active development with feedback from our internal agents team. We believe it is ready to integrate into both academic and commercial pipelines, while also open to further improvement in parallel.   

\section{Conclusion}

We presented Taxi, a system for interpreting agent trajectories at scale, built on a queryable
relational backbone that serves as the integration point for every part of the system. Its main contribution is scalability -- integrated into every design decision, while not excluding potential use of heavier methods when required. Alongside embedding-based clustering and literal querying, the system now supports Jev-style typed decisions for semantic taxonomy routing and evaluation. This extends the same scale-first design to semantic judgments and may become the primary path for them.
We discussed several key points in our design, further elaborated in the appendices. We additionally discussed how we maintain modular structure to make the method suitable for extension and flexible. We hope our system is useful to others building
an interpretive science of agent behavior at production scale, as well as to more modest scales.

\section*{Ethics Statement}

The trajectories analyzed in this work come from the authors' own agentic systems and public data, and do not involve human-subjects data.

\bibliographystyle{plain}
\bibliography{references}

\appendix

\appendixcontents
\section{The Web UI}
\label{app:ui}

The web application is a client of the server described above, not a layer of its own: it is a
single-page React app that reads and writes through the same HTTP API that backs the MCP tool surface, and only provides an easy interface to Taxi for visualization purposes. Below is a description of its components.

\paragraph{Projects sidebar and home screen.} A collapsible sidebar lists the user's own projects
and those shared with them, each with its step count, alongside actions to create a project (name, data source, and for LangSmith sources, an optional run-name filter) and to open the data source
manager. 

\begin{figure}[h]
\centering
\includegraphics[width=\textwidth]{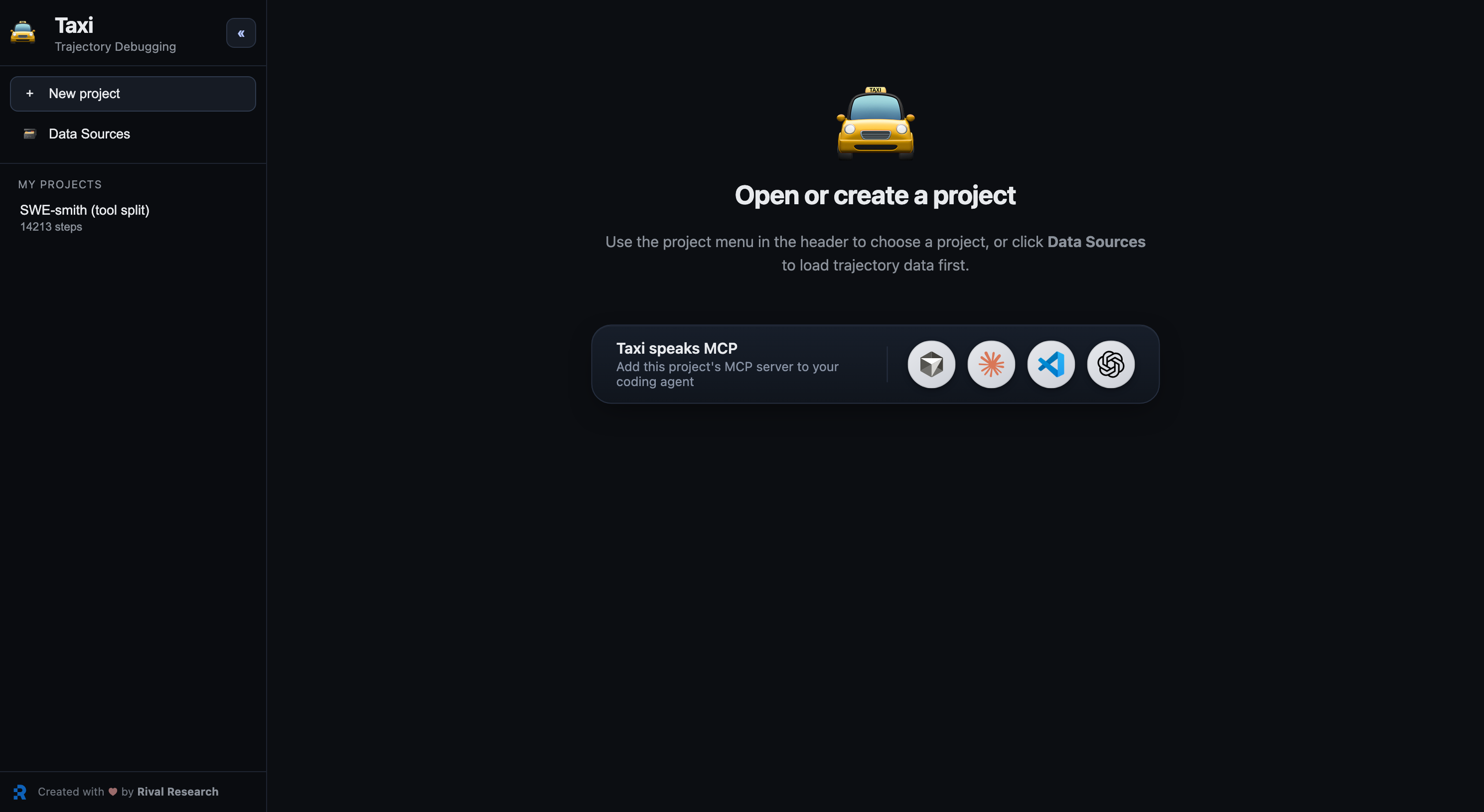}
\caption{Home screen and projects sidebar.}
\label{fig:ui-home}
\end{figure}

\paragraph{Data sources.} Each data source appears as a card with its counts, its current data
version, and controls to sync, embed, or pause and resume a running job. Creating one selects a
loader and fills in that loader's form: a drop zone for a trajectories JSON export; dataset, split,
messages column, row range, and role filter for Hugging Face; workspace, project, metadata-key
filters, and a limit for LangSmith. 

\begin{figure}[h]
\centering
\includegraphics[width=0.9\textwidth]{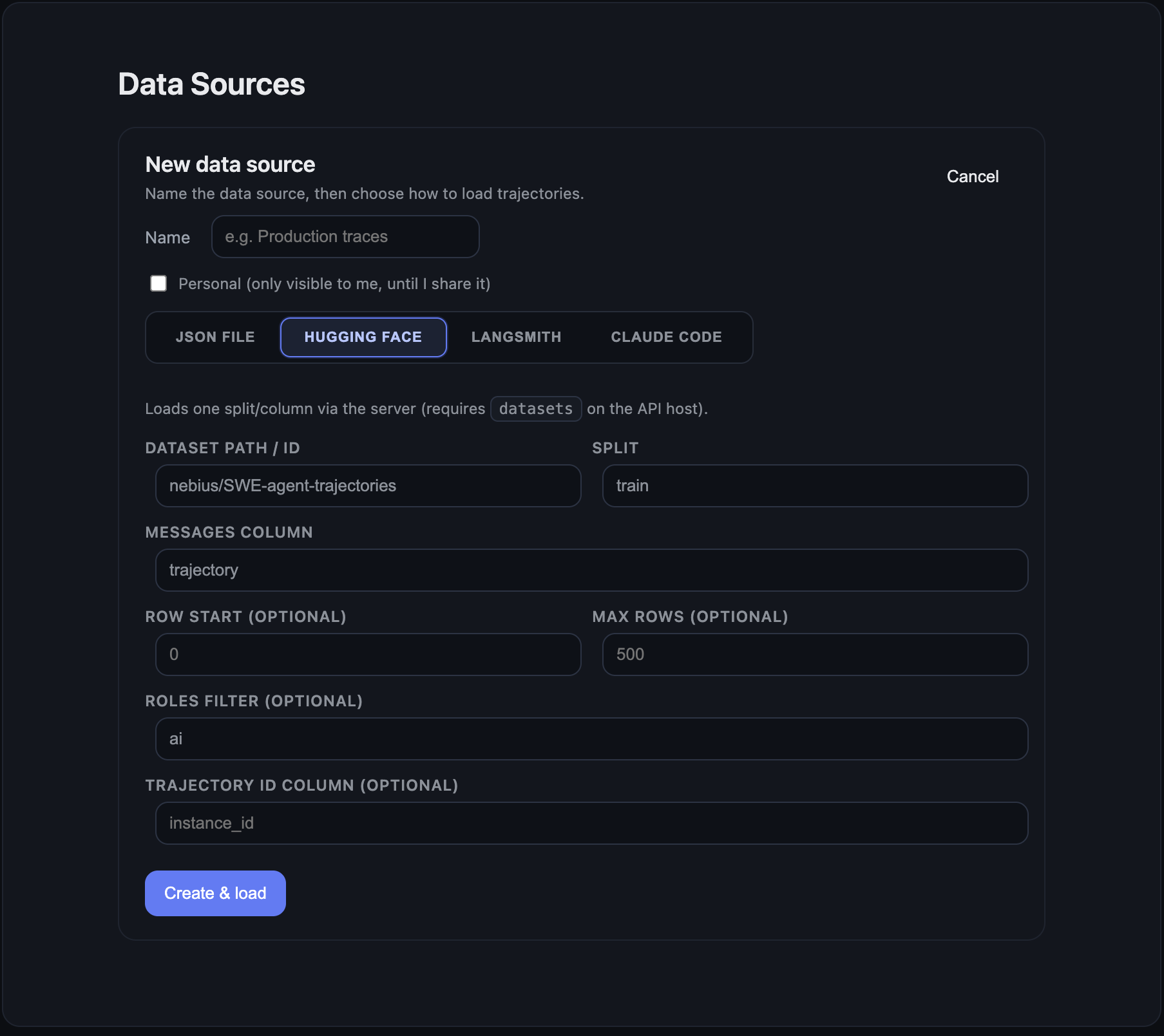}

\caption{Data-source manager and loader forms.}
\label{fig:ui-data-sources}
\end{figure}

\paragraph{Project workspace.} A project opens on three tabs over the same corpus. \emph{Taxonomy}
shows the published tree at its current version -- clusters with their sizes and shares, expandable
into children -- and, upon selecting a cluster, its description, a per-day distribution of its steps
(togglable between absolute counts and share of all steps that day, and draggable to filter to a
date range), and the trajectories it draws from. A sync control sits next to the data-source version
and the version the project has ingested through, so the gap between them is visible before anyone
acts on the tree.

\begin{figure}[h]
\centering
\includegraphics[width=\textwidth]{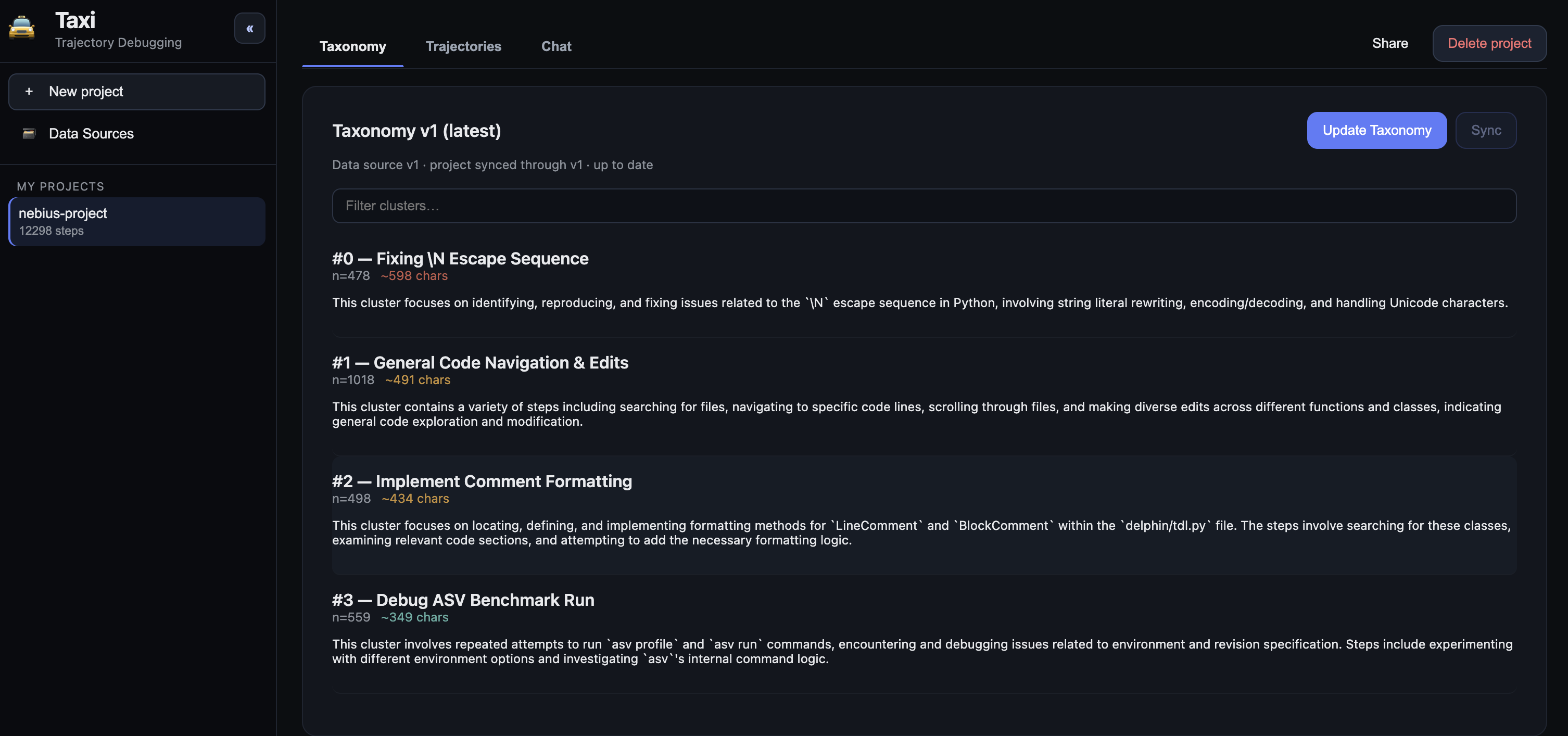}
\caption{Taxonomy tab: published tree and cluster detail.}
\label{fig:ui-taxonomy}
\end{figure}

\paragraph{Trajectories.} The second tab is a paginated list of trajectories with their step counts
and the data version they arrived in; selecting one renders its steps in order, each with its role,
action, and observation, and links back to the run in LangSmith where the source provides one. This
is the view that makes the taxonomy checkable: a cluster is only as good as the steps it turns out
to contain.

\begin{figure}[h]
\centering
\includegraphics[width=\textwidth]{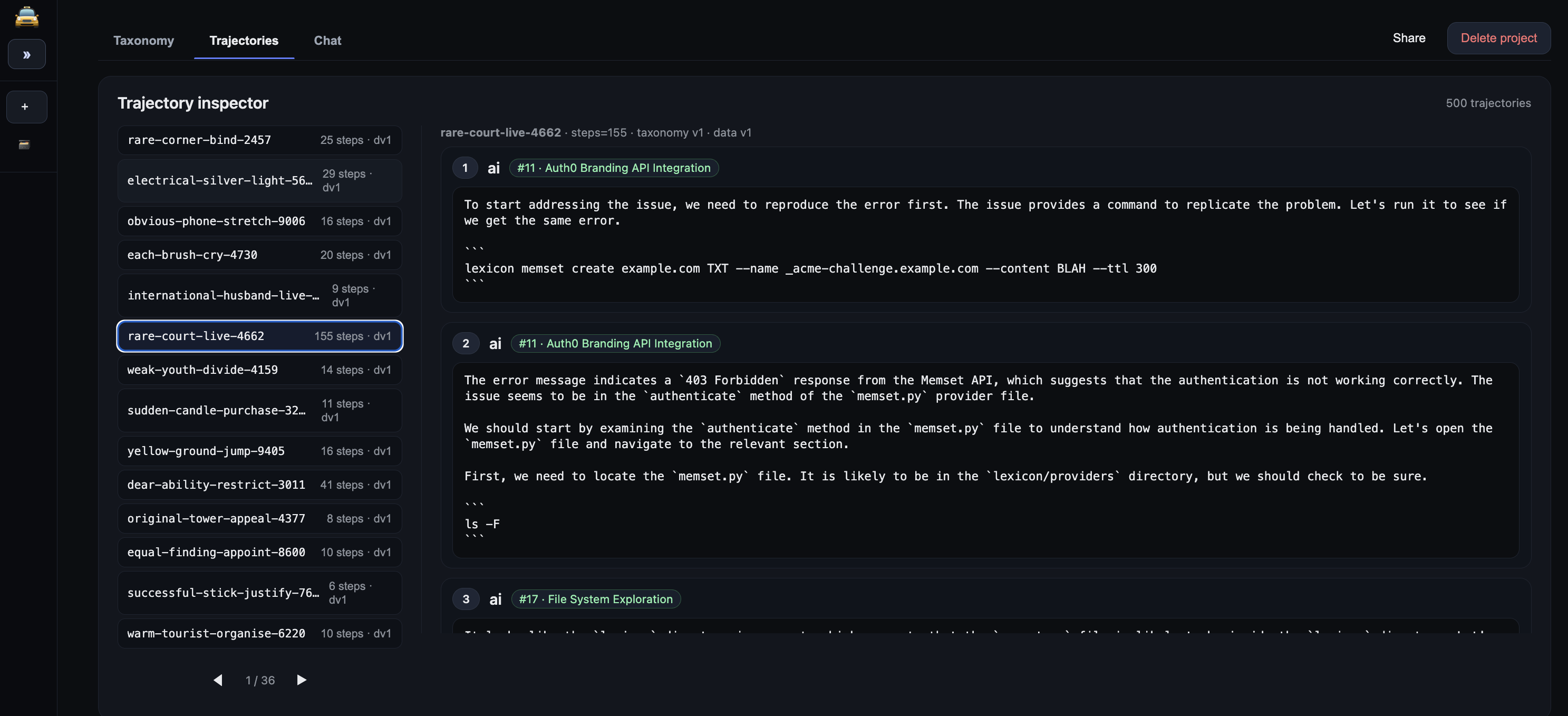}
\caption{Trajectory inspector.}
\label{fig:ui-trajectories}
\end{figure}

\paragraph{Job progress.} Long-running work is not hidden behind a spinner. Wherever a pipeline is
running -- ingest, embedding, clustering, labeling, publishing -- the UI renders its stages in
order, marks which one is live, and shows that stage's progress with a count and an estimate,
driven by the job event stream over server-sent events
; an event log can
be expanded for the raw sequence. The UI holds no progress state of its own: every element here is
a rendering of rows and events in the database, which is why a reload mid-pipeline loses nothing.

\begin{figure}[h]
\centering
\includegraphics[width=\textwidth]{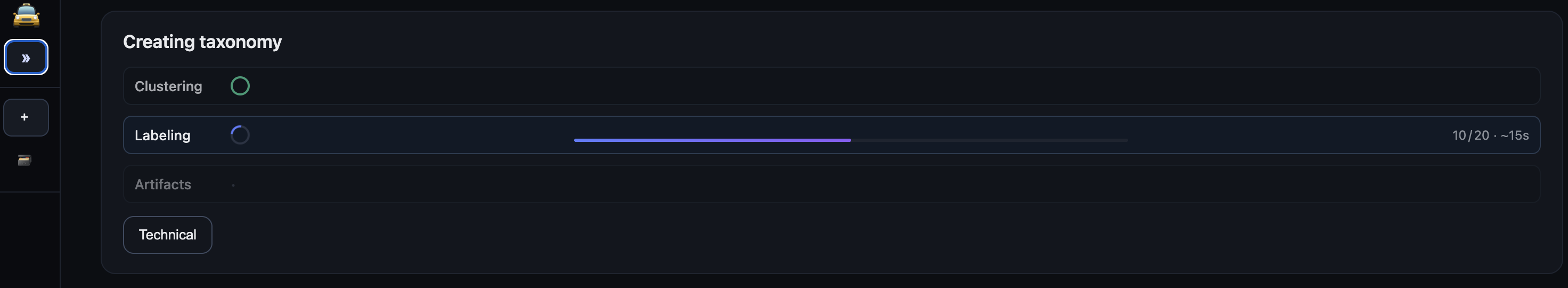}

\caption{Job progress, rendered from the job event stream.}
\label{fig:ui-jobs}
\end{figure}

\section{MCP Tools and Skills}
\label{app:tools_and_skills}

Taxi is designed on top of reusable MCP services and skill documents. Then clients, like the web UI (Appendix~\ref{app:ui}), can be built on top of them, with loose coupling.

Anything a user can do with the web UI -- querying the corpus, refining a taxonomy, publishing an
evaluator, running an insight -- is similarly reachable by a coding agent or any other external caller. That shared surface has two layers:
\begin{itemize}
  \item \emph{MCP tools} -- the primitives: scoped SQL, vocabulary extraction, draft-taxonomy
    operators, and publish/run hooks for evaluators and insights.
  \item \emph{Skills} -- methodology documents that tell an agent \emph{how} to use those
    primitives for a particular job.
\end{itemize}
Capability lives in the documents, not in separate subsystems; adding a new interpretive
procedure means writing a skill over the existing tools rather than growing a privileged code
path the UI alone can see.

\paragraph{Discovery.} A session begins with \texttt{taxi\_init}, which lists every skill under
\texttt{.claude/skills/*/SKILL.md} as
\texttt{\{name, description, whenToUse, resource\}}. The agent reads \texttt{whenToUse} to decide
relevance, then fetches only the matching \texttt{taxi://skills/\{name\}} resource for the full
body. Skills are never duplicated into the server: editing a \texttt{SKILL.md} updates both the
catalog and the body on the next call. The point of this indirection is budget -- an agent loads
methodology only when the task warrants it, rather than carrying every procedure in every turn.

\paragraph{Session scope.} Almost every tool requires an active scope, set before the agent can use them. Scopes are either projects or data sources:
\begin{itemize}
  \item \texttt{set\_project} -- scopes to a project's filters and its latest taxonomy, so the
    pre-filtered views of Appendix~\ref{app:schema}
    (\texttt{project\_trajectories}, \texttt{project\_steps}, \texttt{project\_clusters},
    \texttt{project\_step\_assignments}) apply automatically.
  \item \texttt{set\_data\_sources} -- the lighter path for raw corpora: the same trajectory and
    step views resolve, but the cluster views are empty because there is no taxonomy without a
    project.
\end{itemize}
\texttt{list\_projects}, \texttt{list\_data\_sources}, and \texttt{create\_project} discover or
open the target before either is called.

\paragraph{Primitive surface.} The tools fall into five groups over that scope.

\medskip
\noindent\textit{Scope}
\begin{itemize}
  \item \texttt{taxi\_init} -- list available skills as
    \texttt{\{name, description, whenToUse, resource\}}. Call this before non-trivial work; fetch
    the resource URI only for a skill whose \texttt{whenToUse} matches the task.
  \item \texttt{list\_projects} -- return the caller's projects (id, name). Use before
    \texttt{set\_project} when the target is not already known.
  \item \texttt{set\_project} -- activate one project: its filters and latest taxonomy apply to
    every subsequent query.
  \item \texttt{list\_data\_sources} -- return every data source (id, name, counts, version).
  \item \texttt{set\_data\_sources} -- activate one or more data sources (or \texttt{"*"} for
    all) without a project; cluster views stay empty.
  \item \texttt{create\_project} -- create a project on a named data source and set it as the
    active scope in one step.
\end{itemize}

\noindent\textit{Corpus access}
\begin{itemize}
  \item \texttt{execute\_sql} -- run a read-only \texttt{SELECT} against the database. It is provided access to the scoped
    \texttt{project\_*} views (\texttt{SET TRANSACTION READ ONLY}, capped at 200 rows). This is one of the main workhorses of Taxi, sitting at the base of every textual search into the data.
  \item \texttt{get\_taxonomy\_context} -- return the published cluster tree for the active
    project: each cluster's label, member count, and description. Call at the start of any session
    where taxonomy structure matters.
  \item \texttt{get\_tfidf\_terms} -- top TF-IDF terms from step action text in the active scope.
    The default asks for hundreds of terms because the vocabulary that distinguishes behaviors
    lives in the long tail.
  \item \texttt{sample\_steps} -- draw a random sample of raw step texts, optionally restricted
    to members of a specific cluster.
\end{itemize}

\noindent\textit{Draft taxonomy.}
Construction is session-local until publish: operators mutate an in-memory draft, and nothing is
visible to other clients until \texttt{publish\_taxonomy} commits it.
\begin{itemize}
  \item \texttt{cluster\_by\_embedding} -- run k-means over step embeddings and add the resulting
    clusters to the draft. With \texttt{parent\_id}, only steps that already route to that draft
    cluster are clustered (a scoped split); without it, clustering is at the root. Returns
    representative samples plus cohesion and nearest-neighbor stats for triage. Large scopes are
    subsampled by default for the k-means input; at publish, every step is still routed by nearest
    centroid.
  \item \texttt{cluster\_by\_regex} -- add regex clusters at root or under a parent. Each cluster
    is a pattern matched case-insensitively against step action text; at each tree level, the
    first matching sibling wins. 
  \item \texttt{cluster\_by\_jev} -- add semantic categories defined by a label and description
    at root or under a parent. The configured Jev-style decision model receives the step action
    as state and one Choice question over the sibling definitions, plus ``None of these.'' A
    low-confidence or none-of-these answer leaves the step at its parent. The tool previews the
    resulting assignments on a trajectory sample before publish.
  \item \texttt{construct\_by\_jev} -- (experimental) discover a Jev level, or optionally a
    tree, at root or under a parent, as a background job (Appendix~\ref{app:jev-construction}).
  \item \texttt{apply\_jev\_construction} -- report that job's progress and, once it completes,
    write the constructed categories into the draft.
  \item \texttt{get\_draft\_taxonomy} -- show the current draft tree (labels, patterns, parent
    links) before further edits or publish.
  \item \texttt{clear\_draft} -- discard the draft entirely.
  \item \texttt{draft\_from\_taxonomy} -- copy a published taxonomy version into the draft for
    redesign.
  \item \texttt{add\_draft\_cluster} -- insert one cluster without replacing its siblings. A
    pattern creates a regex cluster; without one, it follows the sibling level's routing kind.
  \item \texttt{remove\_draft\_cluster} -- delete a cluster and all of its descendants.
  \item \texttt{update\_draft\_cluster} -- set or change a cluster's label and description
    (the description defines what the steps do, not how the cluster was produced).
  \item \texttt{split\_draft\_cluster} -- replace one cluster with regex children when the split
    is cleanly describable by patterns, or Jev children when it is cleanly describable by semantic
    definitions; for an embedding split, call \texttt{cluster\_by\_embedding} with that cluster as
    \texttt{parent\_id} instead.
  \item \texttt{merge\_draft\_clusters} -- merge sibling clusters into one (embedding centroids
    combine by size-weighted mean, regex patterns are OR-combined, and Jev descriptions are
    combined into a new semantic definition). Use after
    \texttt{find\_similar\_clusters} flags near-duplicates.
  \item \texttt{find\_similar\_clusters} -- report pairs of embedding clusters whose centroids
    are very close (default cosine $\ge 0.9$), including across different branches -- the signal
    that one theme has fragmented.
  \item \texttt{sample\_cluster\_representatives} -- sample members of a draft cluster in one of
    three modes: \texttt{centroid} (closest to the centroid), \texttt{edge} (farthest from the
    centroid), or \texttt{diverse}.
  \item \texttt{cluster\_terms} -- distinctive TF-IDF terms for one draft cluster versus the
    whole corpus; an aid for naming and for spotting when two clusters are the same theme.
  \item \texttt{publish\_taxonomy} -- commit the draft atomically as a new taxonomy version,
    assign all project steps by tree-walk routing, clear the draft, and return a job id whose
    assignment runs in the background. 
\end{itemize}
The construction loop these tools support -- propose, examine, revise, name as a set, publish --
is the taxonomizer skill's subject (Appendix~\ref{app:taxonomizer}).

\medskip
\noindent\textit{Evaluators} (Appendix~\ref{app:eval-yaml})
\begin{itemize}
  \item \texttt{publish\_eval} -- validate a complete evaluator YAML document and upsert it into
    Postgres by \texttt{name} (re-publishing the same name replaces it in place). No project
    scope required; the YAML itself declares project or data-source scope.
  \item \texttt{run\_eval} -- run every published evaluator, or one by name, optionally narrowed
    to a recency window (\texttt{7d}, \texttt{24h}, or a date). Returns one structured record per
    evaluator (status, filtered/success/failure counts, discrimination-check detail).
\end{itemize}

\noindent\textit{Insights} (Appendix~\ref{app:insights})
\begin{itemize}
  \item \texttt{publish\_insight} -- validate and upsert an insight definition YAML by name
    (data sources to cover, optional Slack channel, free-form focus instructions).
  \item \texttt{list\_insights} -- list enabled published insight definitions.
  \item \texttt{get\_insight\_context} -- after \texttt{set\_data\_sources} on exactly one
    source, return and activate that source's analysis window: a bootstrap week on a first run,
    everything since the last saved report otherwise, or an up-to-date verdict that means skip.
    Scoped views then return only that window.
  \item \texttt{save\_insight\_report} -- persist one (insight, data source, version) Markdown
    report plus a short summary a digest can quote. Idempotent: the first save for that triple
    wins.
  \item \texttt{post\_insight\_reports} -- take every report id saved in the current run and post
    exactly one consolidated Slack digest; full per-source reports stay in Taxi.
\end{itemize}

\paragraph{Skills.} Five skill documents currently sit on that surface. Each is a system prompt
for one job: what to call, in what order, what not to invent, and how to present the result. The
same MCP tools serve all of them.
\begin{itemize}
  \item \texttt{taxi-agent} -- interactive exploration and debugging in the web chat (and similar
    harnesses). Uses \texttt{execute\_sql} and \texttt{get\_taxonomy\_context} only; answers must
    be grounded in query results.
  \item \texttt{taxonomizer} -- agentic taxonomy construction
    (Section~\ref{sec:taxonomizer}). Drives the full draft surface: orient on the corpus (counts,
    TF-IDF, raw samples), propose partitions, examine against evidence, revise (split / merge /
    deepen), name as a set, then publish. Detailed in Appendix~\ref{app:taxonomizer}.
  \item \texttt{taxi-eval} -- author, validate, publish, and run YAML evaluators. Encodes how to
    turn a bug fix's diff into an observable signature, choose between SQL and a Jev-style semantic
    criterion, and decide when to prefer data-source scope over project scope. Detailed in
    Appendix~\ref{app:eval-yaml}.
  \item \texttt{taxi-insights} -- scheduled insight runner in two modes: investigate exactly one
    data source and \texttt{save\_insight\_report}, or compose a digest from that run's saved
    summaries and \texttt{post\_insight\_reports}. Refuses local file or schema repair; stops on
    a required MCP failure. Detailed in Appendix~\ref{app:insights}.
  \item \texttt{slack-agent} -- system prompt for the Slack bot that answers @mentions. Each
    message is a fresh invocation, so it must \texttt{set\_project} every turn; uses the same
    SQL/taxonomy tools as \texttt{taxi-agent}; must end with a \texttt{final\_answer}; formats
    in Slack mrkdwn (no UI link syntax).
\end{itemize}

\section{Taxonomy Construction in Detail}
\label{app:taxonomizer}

A taxonomy is a named tree of step clusters -- one lens on a trajectory corpus, not a replacement
for SQL over the same steps (Section~\ref{sec:backbone}). It serves two jobs at once: a birdseye
overview of what agents actually do, and a stepping stone that gives a human or an agent enough
grasp of the dataset's shape to write targeted queries afterward. Both construction paths below
produce the same artifact shape (a versioned tree in Postgres); they differ in \emph{where}
summarization of ``what behaviors exist'' happens, and in how much human or agent judgment sits
between a clustering call and a published label.

\paragraph{Shared Taxonomy Tree.} Embedding-based, regex-based, and Jev-based clusters can sit in
the same tree, including as siblings of different parents, but at any one level all siblings share
a single routing method.
\begin{itemize}
  \item \emph{Embedding clusters.} k-means over step embeddings proposes a partition; the geometry
    itself is the summarization of the corpus. Coverage at an embedding level is exhaustive among
    steps that have embeddings: every such step is assigned to the nearest centroid. Large scopes
    may subsample to compute centroids (reported explicitly in the tool response); at publish,
    every embedded step is still routed by nearest centroid regardless.
  \item \emph{Regex clusters.} Each cluster is a case-insensitive pattern over step action text.
    Summarization is a separate, upstream step -- currently TF-IDF over the corpus
    (\texttt{get\_tfidf\_terms}), with room for other vocabulary methods -- after which an agent
    (or a human) designs the patterns. Coverage need not be exhaustive: a step matching no sibling
    pattern is simply left unclustered at that level, and that is fine.
  \item \emph{Jev clusters.} Each sibling is defined by a short label and a natural-language
    description. The configured Jev-style decision model receives the step's action text and a
    Choice question whose options are those sibling definitions plus ``None of these.'' The chosen
    sibling wins only when the returned confidence clears a configurable threshold. Thus these
    levels express semantic categories directly, without requiring a centroid or a lexical pattern.
\end{itemize}
The asymmetry matters for method choice. Much of what an agent emits is templated -- tool names,
argument shapes, stereotyped observations, recurring error strings -- and for that material a
literal pattern names a behavior exactly where a semantic neighborhood only approximates it.
Embedding proximity is useful when the categories must first be discovered from the corpus. Once
categories can be stated as definitions, Jev routing can assign variable wording directly against
those definitions. The taxonomizer is free to use any operator per region of the tree rather than
once for the whole corpus. In particular, a regex parent can cheaply pre-filter a Jev child level,
so only steps that reach that branch require a model decision.

\paragraph{Routing.} After publish, every step is assigned by walking the tree from the root:
\begin{itemize}
  \item Regex siblings: first matching pattern wins (order among siblings is significant).
  \item Embedding siblings: nearest centroid wins.
  \item Jev siblings: the model chooses among labels and descriptions; ``None of these'' or a
    low-confidence answer produces no match at that level.
  \item A step that matches a parent but no child stays at the parent.
  \item A step matching nothing at the root is unassigned.
\end{itemize}
 The same tree-walk is what later sync uses: new
steps are routed through the published version without rebuilding it
(Section~\ref{sec:backbone}).

\paragraph{Simple flat clustering.} The default non-agentic pipeline is one k-means pass over the
project's embedded steps, followed by an LLM label for each cluster from a small sample near its
centroid, then an atomic publish. There is no examination loop between proposal and name -- the
clustering call \emph{is} the partition, and each label is written in isolation, blind to sibling
clusters. Despite that myopia it is often useful in practice; most of the production cases in
Appendix~\ref{app:case-studies} came from flat k-means taxonomies. It remains available as a first
pass and as a baseline the agentic path is meant to improve on.

\paragraph{Agentic construction.} The taxonomizer skill (Appendix~\ref{app:tools_and_skills})
treats every clustering call as a \emph{proposal to examine}, not a result to accept. There is no
scoring rubric to optimize: the tools return evidence (samples, distinctive terms, cohesion and
nearest-cluster statistics), and the agent decides what is a coherent cluster, what is a grab-bag,
and what is one theme split in two. The loop, summarized in
Figure~\ref{fig:taxonomy-construction}, is:

\begin{enumerate}
  \item \textbf{Orient.} Corpus size via SQL; hundreds of TF-IDF terms (distinguishing vocabulary
    lives well down the ranking); raw step samples. Get a feel for the behaviors present before
    imposing structure. On a redesign, load the published tree first
    (\texttt{get\_taxonomy\_context} / \texttt{draft\_from\_taxonomy}).
  \item \textbf{Propose.} Cluster at a granularity that matches the thematic richness just seen --
    enough clusters to separate real behaviors, not a deliberately small $k$ to patch later. A
    rough root range is often 8--30 depending on corpus size and diversity; $k$ is adjusted
    up (split) and down (merge) as evidence arrives. Embedding, regex, and Jev operators all accept
    a \texttt{parent\_id}, so a propose step can be scoped to one cluster's members. Jev categories
    are most natural once exploration or embedding clusters have revealed definitions worth routing
    against.
  \item \textbf{Examine.} For each cluster, read centroid samples (the core), edge samples
    (farthest from the centroid -- where a grab-bag shows itself), diverse samples, distinctive
    \texttt{cluster\_terms}, and the triage numbers \texttt{cohesion} and \texttt{nearest\_sim}.
    A handful of centroid samples plus top terms is not enough evidence to name or accept from.
  \item \textbf{Revise.} Confirm against the text before acting:
    \begin{itemize}
      \item Low cohesion, or edge samples that read like a different activity than the core
        $\rightarrow$ grab-bag; split with \texttt{cluster\_by\_embedding(parent\_id=\ldots)} or,
        if the split is cleanly describable, \texttt{split\_draft\_cluster} with regex patterns or
        Jev descriptions.
      \item High \texttt{nearest\_sim}, or \texttt{find\_similar\_clusters} pairing two clusters
        whose terms and samples overlap $\rightarrow$ one theme fragmented;
        \texttt{merge\_draft\_clusters}.
      \item A dominant cluster holding a large share of the corpus $\rightarrow$ usually several
        behaviors collapsed; go deeper.
      \item An empty cluster, or a child that covers almost all of its parent $\rightarrow$ remove
        it / do not nest there.
    \end{itemize}
    Sub-cluster only where distinct sub-behaviors earn a deeper level, not to hit a depth.
  \item \textbf{Name as a set.} Settle membership before naming. Name from centroid samples and
    distinctive terms, never from a handful of stray examples. Then read all names together:
    consistent abstraction level, parallel form, clear contrast. Prefer behavioral names
    (``Searching AWS for Lambda functions'') over structural ones (``step contains serverless
    JSON''). The \texttt{description} stored on a cluster is a 1--3 sentence definition of what
    those steps are or do -- never a process log (``split from cluster~X''). For a Jev cluster,
    this text is also the routing criterion and must be sufficient to decide membership.
  \item \textbf{Publish.} \texttt{publish\_taxonomy} commits the draft atomically as a new
    taxonomy version, assigns all project steps by the tree-walk above,
    and clears the draft. Before calling it, present the proposed tree -- name, size, routing rule
    per cluster -- and say explicitly what was left broad, dropped, or unembedded. Prefer clean
    coverage over forced coverage: a smaller set of strong, distinct clusters beats padding the
    tree to cover every last step.
\end{enumerate}

\paragraph{Draft vs.\ published.} While the loop runs, the tree is session-local draft state
(Appendix~\ref{app:tools_and_skills}). Other clients and the project's published taxonomy are
untouched. Publish is the only transition that makes the tree durable, and it is all-or-nothing:
version row, cluster rows, and step assignments land in one Postgres transaction, or none do
(Appendix~\ref{app:schema}). Prior versions remain readable; sync of newly ingested steps always
targets the version the project currently uses.

\paragraph{Relation to the flat pipeline.} The agentic path reuses the same operators and the same
publish/routing machinery as the flat pipeline. What it adds is everything between propose and
name: scoped re-clustering, examination against evidence, merge and split, and naming with
visibility into sibling clusters. The flat pipeline is the propose-and-name fragment of
Figure~\ref{fig:taxonomy-construction} with the middle boxes removed; the taxonomizer skill is
the full loop, written as methodology over the MCP surface rather than as a second clustering
subsystem.

\section{Experimental: Jev-Based Taxonomy Construction}
\label{app:jev-construction}

\begin{figure}[h]
\centering
\includegraphics[width=0.95\linewidth]{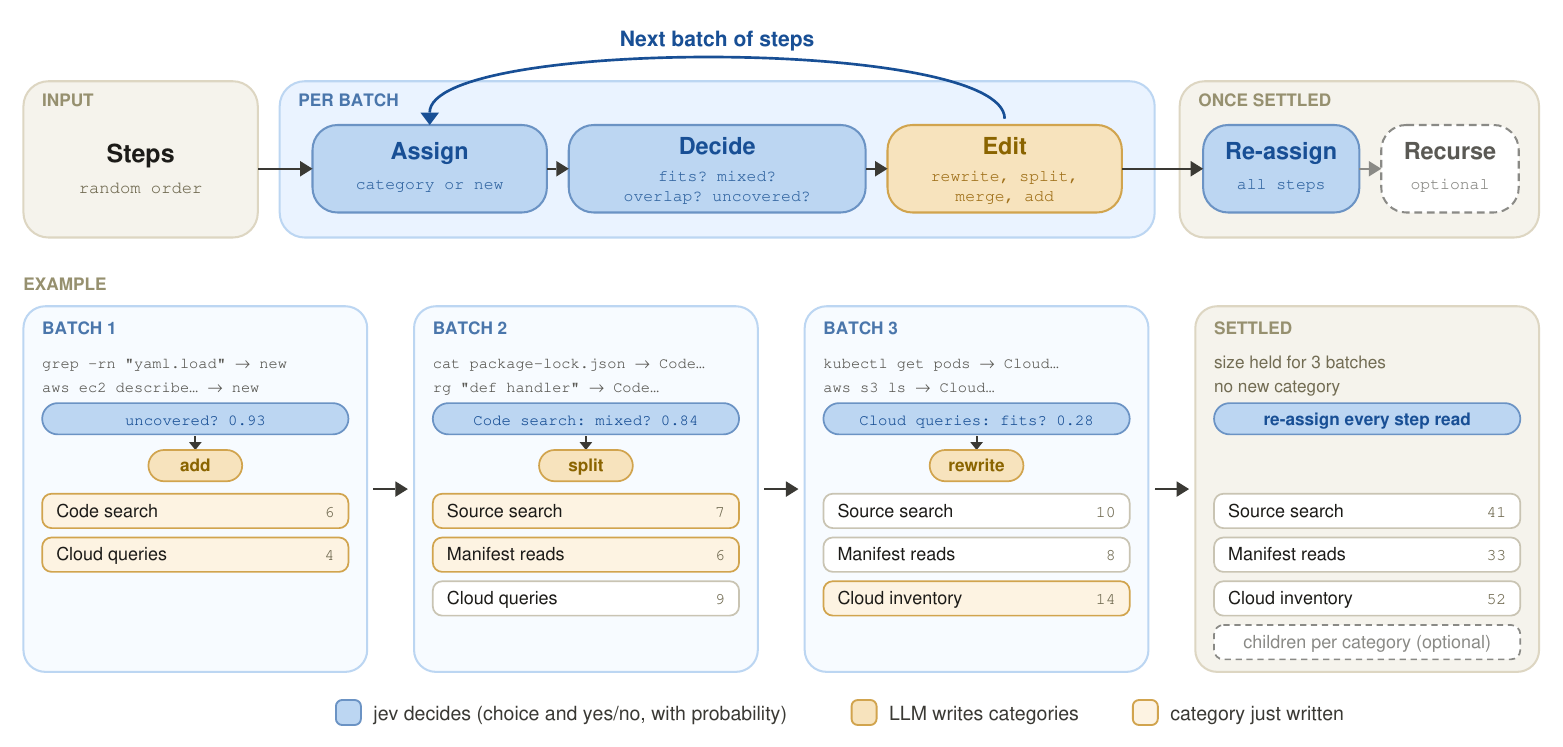}
\caption{Jev-based construction of one taxonomy level. Each batch is assigned, judged, and
edited; once the level settles, every step is re-assigned to the finished categories.}
\label{fig:jev-construction}
\end{figure}

Jev routing (Appendix~\ref{app:taxonomizer}) assigns steps to categories that are already
defined. \texttt{construct\_by\_jev} discovers those categories from the corpus, the way
\texttt{cluster\_by\_embedding} discovers embedding clusters. The division of labor is strict:
the decision model makes every decision -- which category a step belongs to, and whether the
level should change -- and an LLM is called only to write a category once a change has been
decided. Figure~\ref{fig:jev-construction} summarizes the procedure.

\paragraph{Online construction.} The scope's steps stream in a fixed random order, a batch at a
time. Each new step is assigned once, with the same question routing uses, to one of the
current categories or to ``new behavior.'' After each batch, the decision model answers yes/no
questions over all the steps seen so far:
\begin{itemize}
  \item \emph{Fits?} Does a category's description fit its member steps?
  \item \emph{Mixed?} Does a category hold more than one behavior?
  \item \emph{Overlap?} Do the members of two categories fit both equally well? Candidate pairs
    come from asking which other category is closest to each category, and from how often two
    categories competed for the same steps.
  \item \emph{Uncovered?} Do the uncategorized steps share a behavior no category covers?
\end{itemize}
The answers' probabilities rank the candidates, and only the top few edits per batch are made.
An LLM then writes each one: a rewritten description, the parts of a split, a merged category, or
a new category. The LLM sees the other categories, so it does not recreate one that exists. A
split is kept only if the decision model can tell its parts apart, and only the steps an edit
touched are assigned again.

\paragraph{No target count.} The number of categories is not fixed in advance. Two share bounds,
relative to the steps seen, frame the level: a category above the upper bound is split ahead of
others, and no merge may create one; a category below the lower bound is dropped. Within those
bounds, the count is whatever the decisions above produce. An optional ceiling is available.

\paragraph{Stopping and the final pass.} The level is \emph{settled} once its size has held for a
few batches and the last batch created no category, so construction usually reads only part of
the sample. Every step read is then assigned once more to the finished categories, so the sizes
and examples reported are what publishing will produce; a category left empty is dropped.

\paragraph{Optional hierarchy.} With a depth above one, every category with enough member steps
gets children, constructed the same way over its own members, with share bounds relative to its
own steps. A node that keeps fewer than two children stays a leaf, so depth and width come from
the data. The result is written into the draft as ordinary Jev clusters, routed and published
like any other level. Construction runs as a background job and checkpoints after every batch,
so an interrupted job resumes where it stopped.

\section{Evaluators}
\label{app:eval-yaml}

An evaluator (Section~\ref{sec:ecosystem}) is a single YAML document. Once published it is stored
verbatim in the \texttt{evaluators} table (Appendix~\ref{app:schema}) and can be run from any
client without a local checkout. The full field set:

\begin{lstlisting}[language=yaml]
name: required-tool-called       # eval name
project: my-project              # option A: the scope of the eval is a project

# data_source: my-ds             # option B: a single data source
# data_sources: [ds-a, ds-b]     # option C: several (union)
# data_sources: "*"              # option D: every data source

tickets: TAXI-123                # optional; string or list
# metadata:                      # optional freeform map
#   owner: platform

description: >                   # for human readability: what it guards against
  The tool required_tool() must be used in every trajectory of the cloud agent.

failure_examples:                # sanity check: trajectory ids known to fail
  - past-bad-run-1
  - past-bad-run-2

filter_sql: |                    # optional: subset of relevant trajectories
  SELECT id FROM project_trajectories
  WHERE meta->>'run_name' = 'cloud-agent'

success_sql: |                   # what is considered a good trajectory
  SELECT DISTINCT s.trajectory_id AS id
  FROM project_steps s
  WHERE s.tool_calls::text ~* 'required_tool'

# or alternatively, use failure_sql (what is considered a bad trajectory)

# A semantic criterion can instead use the configured Jev-style model:
# failure:
#   jev: Does the agent stop without attempting a fix?
#   threshold: 0.7             # hit when P(yes) >= threshold; default 0.5
#   per: trajectory            # or step
#   state: |                   # optional SQL returning (id, state)
#     SELECT trajectory_id AS id, action AS state FROM project_steps

\end{lstlisting}

Exactly one of \texttt{project} or \texttt{data\_source}/\texttt{data\_sources} is required, and
the choice decides what the evaluator can see. A project scope reads the filtered views of
Section~\ref{sec:backbone}, so the project's own filters and its latest taxonomy apply and
\texttt{project\_clusters} and \texttt{project\_step\_assignments} are meaningful; a data-source
scope queries the underlying trajectory and step tables, where those two cluster views are empty.

Evaluation then proceeds in three stages, each narrowing the previous: scope, then filter, then
criterion. If \texttt{filter\_sql} is set, the scope is intersected with the ids it returns, and an
empty filtered set yields skip rather than pass, since nothing was actually tested. The filtered set
is intersected with the criterion, of which exactly one form may be given. SQL criteria retain the
existing behavior: with \texttt{success\_sql} the evaluator passes iff every filtered trajectory is
a hit, and with \texttt{failure\_sql} it passes iff none is. A \texttt{success} or \texttt{failure}
mapping may instead contain a \texttt{jev} yes/no question. By default the decision model sees all
actions in a trajectory in order; with \texttt{per: step}, it sees one action at a time and the
trajectory is a hit when any step reaches the probability threshold. An optional \texttt{state}
query can select other text or structured state while preserving the same trajectory identifiers.
SQL filters and state queries therefore continue to perform cheap exact narrowing before any model
call. Results are reported as \texttt{n\_success}/\texttt{n\_filtered} together with a failure
count. There are four possible outcomes: success, fail, skip, and error. Skip means the filtered set was empty, and error is
reserved for problems with the definition rather than the corpus --- an unresolvable or ambiguous
scope name, SQL that does not execute, or a failed discrimination check. Error is deliberately
distinct from fail because the two demand different responses: a fail is a finding about the agents,
an error is a bug in the check. 

Ids listed under \texttt{failure\_examples}
are verified \emph{before} the live corpus result. Each must exist in scope, match the filter SQL, and be classified as
a failure by the criterion. If any does not hold, the evaluator reports an error even when the current corpus would otherwise
pass.

The evaluator is authored as a YAML document, and a durable copy lives in Postgres. Publishing takes either the complete YAML
text through the \texttt{publish\_eval} tool, validates it, and upserts by
\texttt{name}, so re-publishing the same name replaces that evaluator in place. 
\texttt{run\_eval} executes every published evaluator, or one by name, optionally within a recency
window, and returns one structured record per evaluator carrying its status, its filtered, success
and failure counts, and any error or discrimination-check detail. 
These reports are created as scheduled reports that keep a per-data-source watermark -- the ingest version last
evaluated -- so it considers only trajectories that arrived past it, and advances it once the run
finishes. 

Beyond hand-written invariants, the same format supports turning engineering history into checks, and
that methodology is documented in the evaluator skill document. Candidates come from reading a commit log or
closed tickets broadly. Then the agent is tasked with
translating that into an \emph{observable signature}, i.e., something we can query over trajectories.
\section{Insight Reports}
\label{app:insights}

An insight is a recurring investigation of one or more data sources,
defined in YAML, published to Postgres, and run by an agent on a schedule. A definition names the
insight, the data sources it covers (a list, or \texttt{"*"} for every source), and the channel its
digest is posted to; anything else in the document is carried through to the investigating agent as
context. As with evaluators, publishing upserts by name.

There are three modes covering three cases: if no report already exists for this data source, a bootstrap run runs over last week's trajectories; any other case gets everything from the data version its last report covered, unless there was no new data ingested, which immediately receives an up-to-date verdict and stops. 
Each investigation saves one Markdown report plus a short summary that a Slack digest can quote.

Findings are
ordered by how much they would surprise a competent reviewer, and a corpus's own outcome tallies are
background rather than the headline unless the rate itself is the surprise. Agent \emph{behavior} is
the subject --- strategy, tool choice and sequencing, instruction following, wasted or repeated
action, recovery, stopping behavior --- rather than a catalog of the tasks or artifacts the agents
happened to touch. When a batch is small enough to read end to end it is read end to end rather than
sampled, and every number must come from a query run in that same investigation.

\section{Real-World Case Studies}
\label{app:case-studies}

Below is a short anecdotal description of real-world cases our users found during organic use of our system.

\begin{table}[h]
\centering
\small
\begin{tabular}{@{}p{0.28\linewidth}p{0.32\linewidth}p{0.32\linewidth}@{}}
\toprule
\textbf{Case} & \textbf{Surfacing cluster} & \textbf{Underlying issue} \\
\midrule
Broken tool regex & a distinct ``search returned nothing'' cluster & a valid pattern was silently
reported as a non-match by a downstream tool \\
Scope violation & a cluster of steps outside the declared task scope & the agent expanded its
investigation beyond an explicitly user-scoped target \\
Redundant lookup & a repeated ``re-derive static infrastructure fact'' cluster & the same
unchanging information was rediscovered from scratch on every run \\
Missing dependency & a repeated ``tool not found, retry, abandon'' cluster & the agent assumed a
sandbox capability that was never actually installed \\
Expired credential & a cluster of near-identical authentication failures & a revoked API token was
causing failures across many unrelated runs \\
\bottomrule
\end{tabular}
\caption{Five production cases in which taxonomy clustering, rather than an outcome label,
surfaced a behavioral failure that would otherwise have gone unnoticed.}
\label{tab:cases}
\end{table}

\textbf{Broken tool regex.} A cluster of steps whose observations all resembled a failed search
turned out, upon inspection, to share a specific and surprising cause: a downstream tool's
pattern matching silently treated regex patterns as a non-match, so the agent
was truthfully told ``no matches found'' for searches that should have succeeded. No individual run
looked like a bug from the outside -- the agent adapted, tried alternatives, and often still
completed its task -- so nothing about task outcome flagged the issue. The cluster made the
\emph{frequency and uniformity} of this specific failure shape visible in a way no single trace
could.

\textbf{Scope violation.} A cluster grouped steps whose content fell outside a task's explicitly
stated boundary (investigating only a designated, user-controlled input) instead of a
generalized exploration the agent was never authorized to perform. Individually, each such run
still ``worked'' by task-completion standards; the cluster is what made the systematic boundary
violation legible across many otherwise-unrelated runs.

\textbf{Redundant lookup.} A cluster captured many independent runs each re-deriving the same
essentially static piece of infrastructure information from scratch, rather than treating it as
known. None of these runs failed, and none was individually expensive -- the cost was in the
aggregate, thousands of repeated, avoidable steps, visible only once grouped.

\textbf{Missing dependency.} A cluster of steps shared a distinctive failure-then-abandon shape:
the agent assumed a tool was available in its execution environment, attempted to use it
repeatedly, failed each time, and eventually gave up on that line of work. The cluster's
consistency across runs pointed directly at an environment configuration gap rather than a
per-task modeling failure.

\textbf{Expired credential.} A cluster of near-identical authentication failures, all citing the
same external service, made a revoked API credential visible as a single incident affecting many
runs simultaneously, rather than as scattered, seemingly unrelated failures that might otherwise
have been investigated (and worked around) independently, run by run.

\section{Taxi on SastBench}
\label{app:sastbench}

This appendix walks through one corpus with Taxi, layer by layer: querying, vocabulary, taxonomies,
state machines, decision-model questions, and analyses built on top of the discretized runs. Each part
shows what a capability surfaced, and every claim was checked against its obvious confounds before
being kept.

\paragraph{Setup.} Four agents --- Claude Sonnet 4.5, Gemini 3 Pro, Gemini 2.5 Pro and DeepSeek R1 ---
triage the same 2,737 static-analysis findings of SastBench~\citep{feiglin2026sastbench} with the same ReAct harness (read, grep,
symbol lookup, a vulnerability-pattern tool, at most 20 steps), deciding whether each finding is a
true or a false positive. That gives 10,515 runs and 87,140 steps, loaded as one corpus with a view
per agent. 89\% of the findings are false positives. Step text is each step's reasoning and tool
call; the tool's output is kept as the step's observation.

\subsection{Querying the corpus}
\label{app:sastbench-sql}

\textbf{Errors are shared, not idiosyncratic.} Grouping runs by finding in SQL, 574 findings are
judged wrong by all four agents, against about 120 if each agent erred independently at its own rate.

\textbf{One repository behind half of the shared errors.} A text search over the file paths the agents
read places 265 of those 574 findings (46\%) in a single PHP repository, which contributes 803 of the
2,737 findings across 42 commits. Its findings are labeled true positive 5\% of the time, yet the
agents call 58\% of them vulnerable, and the code their verdicts cite reads as genuinely exploitable
(request parameters concatenated into SQL, class names instantiated from request input). Either
labels in that repository deserve another look, or the benchmark labels issues that predate a commit
as false positives. Excluding it, errors remain shared: 309 findings wrong for all four agents
against about 65 expected (4.8$\times$).

\textbf{Tool-failure profiles.} Pattern matching over observations separates the agents: Claude's
failures are mostly empty searches (51\%), the Gemini models mostly hit file-size limits (38--39\%)
and missing paths (28--35\%), and Gemini 3 Pro and DeepSeek R1 produce malformed tool calls (303 and
245 steps). After a failure, Claude and Gemini 2.5 Pro usually adjust the same call (55\% and 62\%),
DeepSeek R1 usually switches tools (60\%), and almost none retry the identical call.

\subsection{Vocabulary}
\label{app:sastbench-tfidf}

TF-IDF over each agent's steps, against the other three, gives each agent a recognizable style:

\begin{table}[h]
\centering
\small
\begin{tabular}{@{}p{0.22\linewidth}p{0.72\linewidth}@{}}
\toprule
Agent & Most distinctive vocabulary \\
\midrule
Claude Sonnet 4.5 & the mechanism: \emph{loose comparison, prepared statements, template literals, innerHTML, format strings} \\
DeepSeek R1 & taint: \emph{unsanitized, user-controlled data, flows, should verify, security patterns} \\
Gemini 2.5 Pro & its own process: \emph{first step, final decision, enough information, wants to analyze} \\
Gemini 3 Pro & navigation: \emph{around lines, search usages, suspect} \\
\bottomrule
\end{tabular}
\caption{Distinctive terms per agent (TF-IDF, each agent's steps against the other three).}
\label{tab:sastbench-tfidf}
\end{table}

\subsection{Two taxonomies, two lenses}
\label{app:sastbench-taxonomies}

\textbf{Behavior, from textual regularity.} A regex taxonomy built from the regularity of the text
splits steps by tool, then by what the agent says it is doing (looking for protections, tracing data
flow, reading the flagged code, recovering from an error), what the arguments target (test files,
templates, configuration, third-party code; input sources, sinks, definitions), and how the verdict
is justified (input not attacker-controlled, framework protects, missing sanitization, and so on):
29 leaves, with no embeddings.

\textbf{Subject, from embeddings.} A mixed tree keeps the regex roots by tool and puts k-means
clusters under each, named by an agent from samples and distinctive terms (36 leaves). Its clusters
split by what the code is about (CSRF forms, SQL queries, redirects, container configuration,
vendored front-end libraries) rather than by what the agent does: the same steps, a different lens.

\textbf{Some subjects fail for every agent.} Grouping runs by the subject cluster of their verdict
step (Table~\ref{tab:sastbench-topics}), front-end XSS and integrity findings and network/TLS findings
are judged correctly about a quarter of the time by all four agents, while code identified as
third-party or non-production is judged correctly almost always. The gap is not a few commits (each
subject spans 134--209 commits; dropping its three largest moves accuracy by at most 0.02) and not
the base rate (5--18\% true positives in every subject). Front-end verdicts say ``vulnerable'' 88\%
of the time.

\begin{table}[h]
\centering
\small
\begin{tabular}{@{}lrr@{}}
\toprule
Verdict subject & runs & accuracy \\
\midrule
front-end XSS, integrity (SRI) & 778 & 0.24 \\
network, TLS, cookies & 743 & 0.27 \\
PHP web application & 834 & 0.45 \\
container privileges & 681 & 0.45 \\
injection, deserialization & 1,221 & 0.48 \\
ReDoS, path traversal & 731 & 0.61 \\
third-party library & 875 & 0.87 \\
test, example or unused code & 1,052 & 0.88 \\
\bottomrule
\end{tabular}
\caption{Accuracy by the subject of the verdict step (mixed taxonomy), all agents, excluding the
repository of Appendix~\ref{app:sastbench-sql}.}
\label{tab:sastbench-topics}
\end{table}

\subsection{State machines}
\label{app:sastbench-states}

\begin{figure}[p]
\centering
\includegraphics[width=\linewidth,height=0.85\textheight,keepaspectratio]{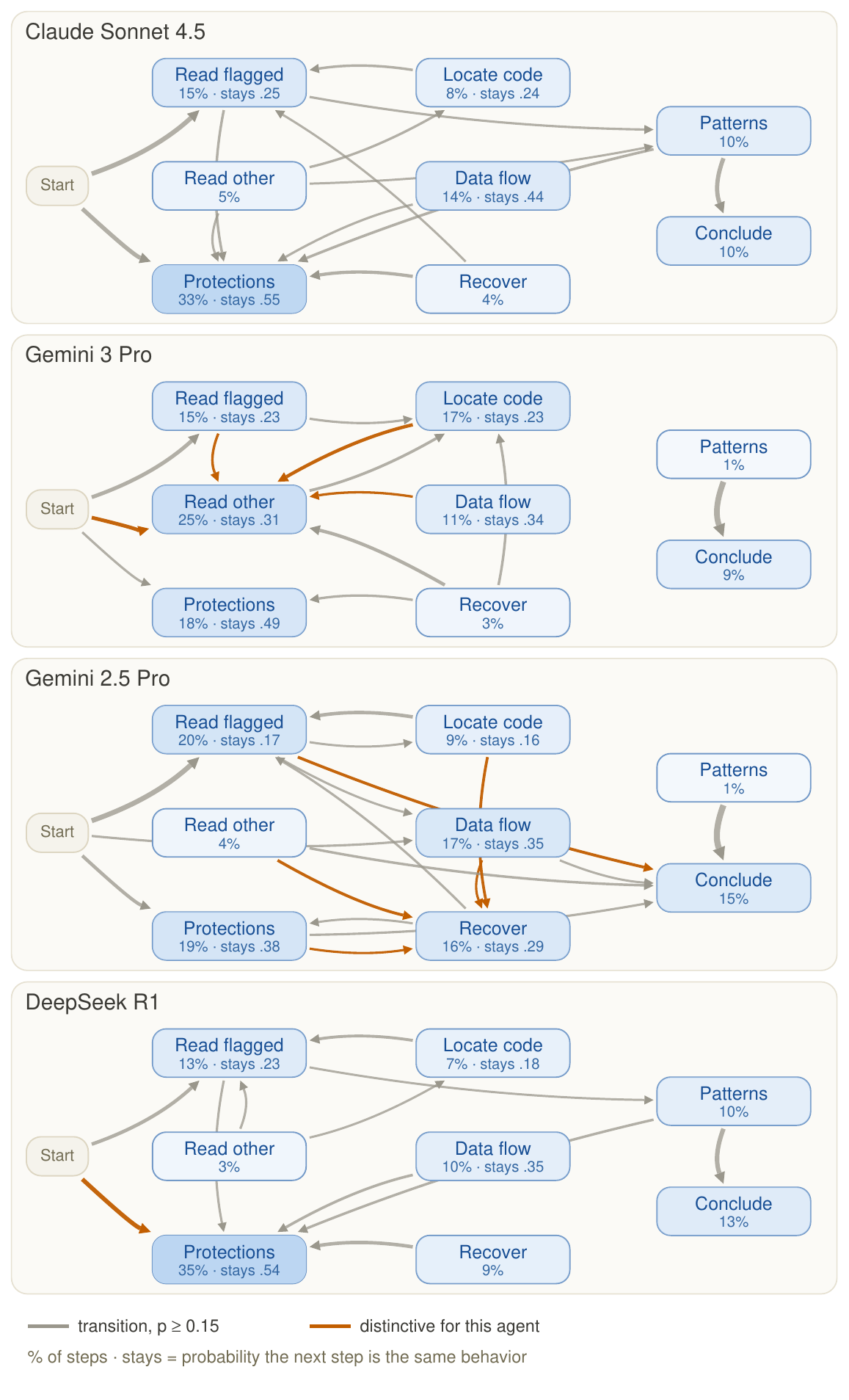}
\caption{State machines of the four agents over eight behaviors. Each box gives the share of the
agent's steps in that behavior (also its shade) and, where above 0.15, the probability that the next
step stays in it. Edges are transitions with probability at least 0.15, width by probability; orange
ones are at least 0.12 more likely for that agent than for the other three.}
\label{fig:sastbench-states}
\end{figure}

\begin{figure}[p]
\centering
\includegraphics[width=\linewidth,height=0.85\textheight,keepaspectratio]{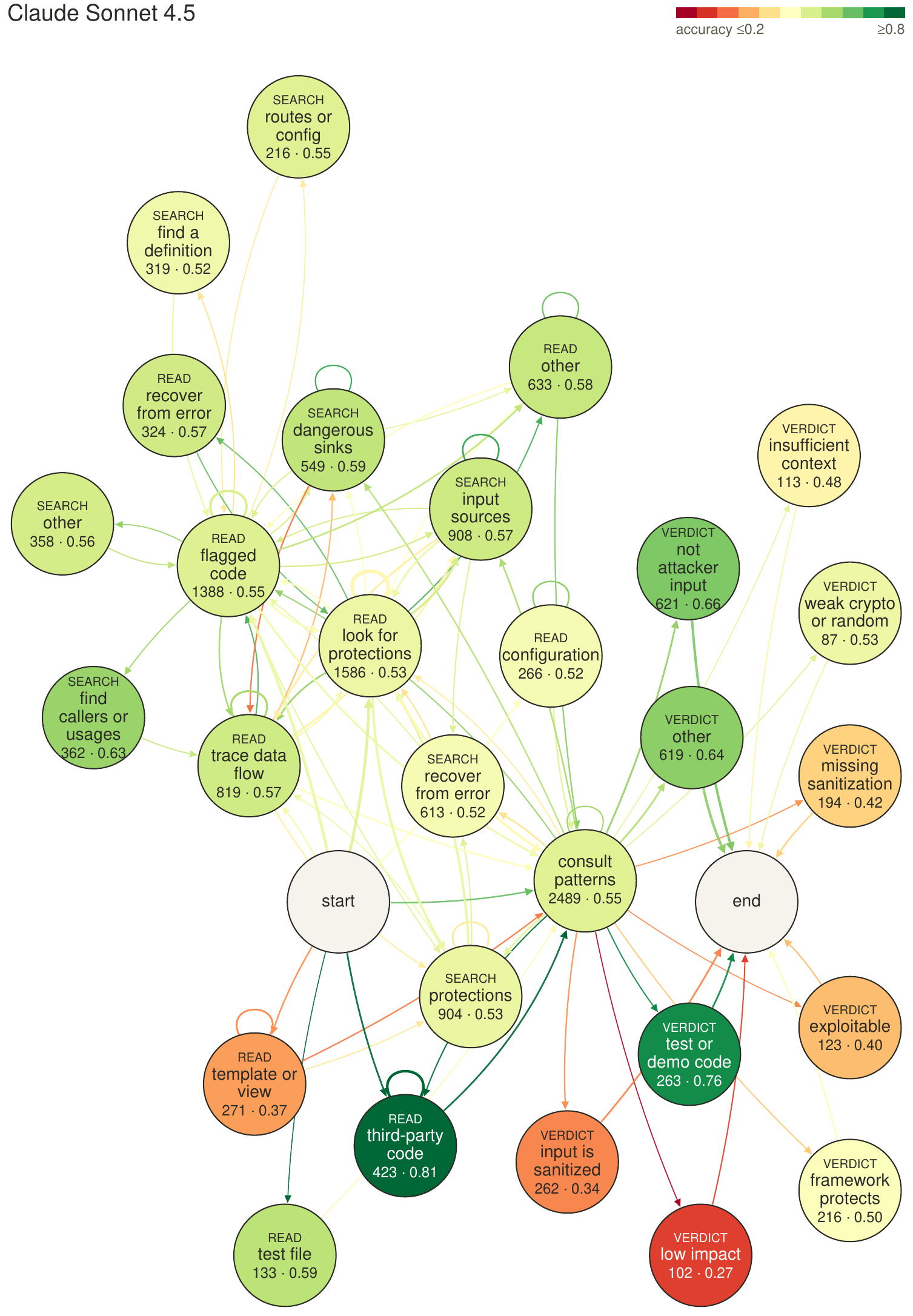}
\caption{Claude Sonnet 4.5 over 29 behaviors. Each circle holds a behavior (with its tool: read, search or verdict), the runs that pass through it and their accuracy, which also sets its color (red to green). Edges are transitions seen at least 80 times plus each behavior's strongest way in and out, colored by the accuracy of the runs that make them, width by count.}
\label{fig:sastbench-flow-claude}
\end{figure}

\begin{figure}[p]
\centering
\includegraphics[width=\linewidth,height=0.85\textheight,keepaspectratio]{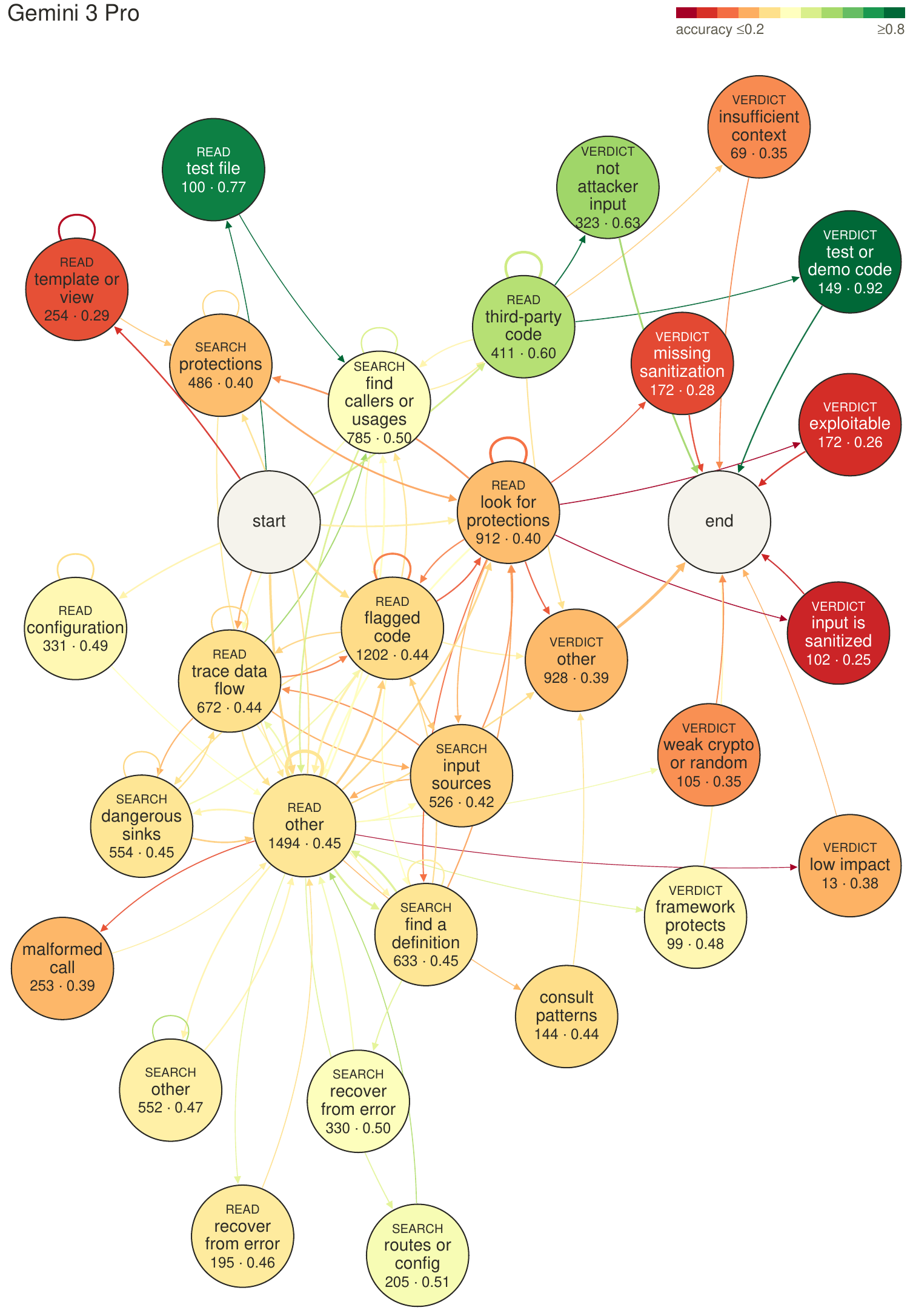}
\caption{Gemini 3 Pro over 29 behaviors, as in Figure~\ref{fig:sastbench-flow-claude}.}
\label{fig:sastbench-flow-gemini3}
\end{figure}

\begin{figure}[p]
\centering
\includegraphics[width=\linewidth,height=0.85\textheight,keepaspectratio]{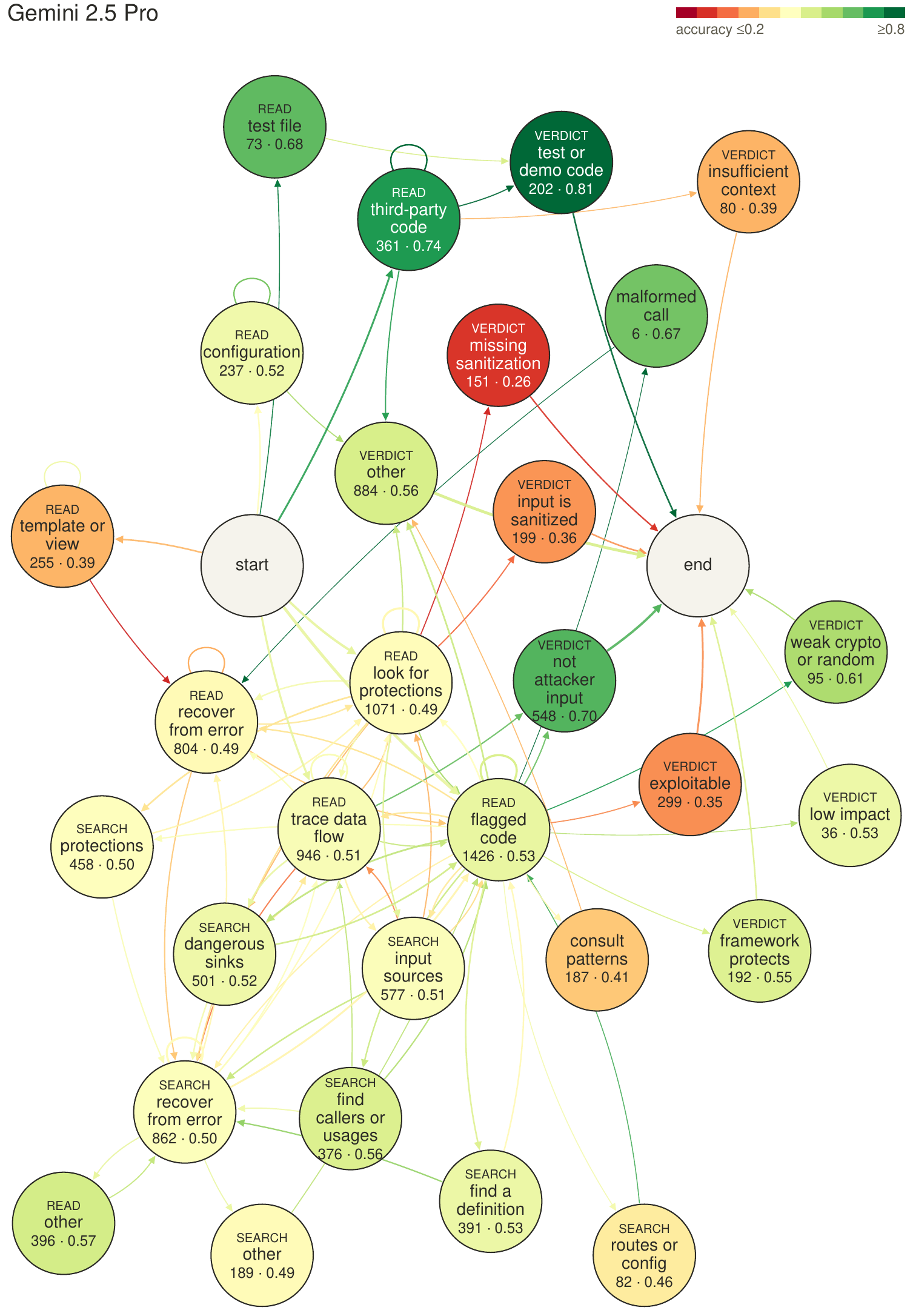}
\caption{Gemini 2.5 Pro over 29 behaviors, as in Figure~\ref{fig:sastbench-flow-claude}.}
\label{fig:sastbench-flow-gemini25}
\end{figure}

\begin{figure}[p]
\centering
\includegraphics[width=\linewidth,height=0.85\textheight,keepaspectratio]{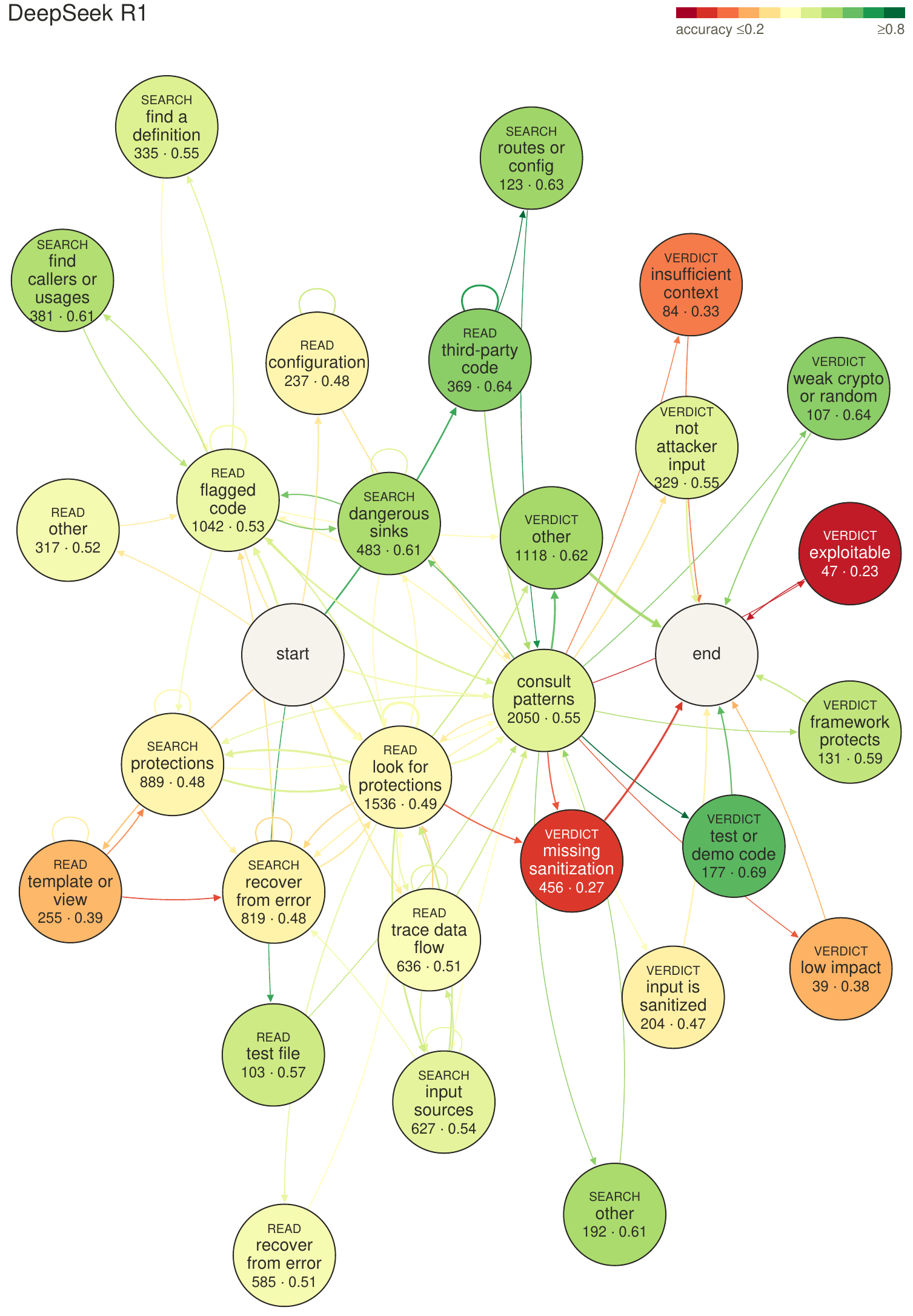}
\caption{DeepSeek R1 over 29 behaviors, as in Figure~\ref{fig:sastbench-flow-claude}.}
\label{fig:sastbench-flow-deepseek}
\end{figure}

Discretized runs read as state machines. Figure~\ref{fig:sastbench-states} shows each agent over
eight behaviors, with the transitions distinctive to that agent highlighted: Gemini 3 Pro loops
between locating code and reading code other than the flagged location; Gemini 2.5 Pro chains error
recovery into more error recovery and often concludes right after reading the flagged code; Claude
and DeepSeek consult the vulnerability-pattern tool just before concluding, which the Gemini models
rarely do. Figures~\ref{fig:sastbench-flow-claude}--\ref{fig:sastbench-flow-deepseek} show the same at
29 behaviors, with every behavior and transition colored by the accuracy of the runs that pass
through it. Drawing only transitions between consecutive behaviors loses little at this granularity:
on held-out runs, knowing the current behavior cuts the uncertainty about the next one from about 4.2
to 3.3 bits per step, and knowing the one before adds at most 0.02.

\subsection{Decision-model questions}
\label{app:sastbench-jev}

Some questions are semantic and polar, which keyword rules cannot capture: a pattern for ``input is
sanitized'' also matches ``\texttt{htmlspecialchars} only prevents XSS, not\ldots''. A Jev-style
decision model answers them directly.

\textbf{Verdicts without an attacker path are half as precise.} Asked of 1,546 sampled ``vulnerable''
verdicts --- does the agent show how attacker-controlled input reaches the flagged code, or only that a
risky pattern is present? --- verdicts that show a path are right 18\% of the time (95\% interval
15--22\%, resampling commits), and those that do not, 9\% (7--12\%). Claude and Gemini 2.5 Pro show a
path in half their ``vulnerable'' verdicts, DeepSeek R1 and Gemini 3 Pro less often (38\% and 43\%).
Even with a path, ``vulnerable'' calls are mostly wrong here.

\textbf{Most investigations are silent until the verdict.} Asked at two points of 1,200 sampled runs
(after the first observation, and 60\% of the way through) whether the reasoning leans toward a
vulnerability, toward a false positive, or neither yet (Table~\ref{tab:sastbench-stance}): Claude
states a position mid-run in half its runs, Gemini 3 Pro in one in six, its ``vulnerable'' verdicts
appearing only at the end. A position, once stated, is the final verdict 90\% of the time.

\begin{table}[h]
\centering
\small
\begin{tabular}{@{}lrrrr@{}}
\toprule
Mid-run stance & vulnerable & false positive & none yet & final ``vulnerable'' \\
\midrule
Claude Sonnet 4.5 & 0.22 & 0.29 & 0.49 & 0.45 \\
Gemini 3 Pro & 0.09 & 0.07 & 0.83 & 0.63 \\
Gemini 2.5 Pro & 0.10 & 0.18 & 0.73 & 0.47 \\
DeepSeek R1 & 0.17 & 0.08 & 0.74 & 0.50 \\
\bottomrule
\end{tabular}
\caption{What the agent's reasoning leans toward 60\% of the way through a run (decision model, 300
runs per agent).}
\label{tab:sastbench-stance}
\end{table}

Both studies together took about 6,300 decision-model calls, a few minutes each at the provider's
rate limit.

\subsection{Beyond the taxonomy}
\label{app:sastbench-beyond}

\textbf{The task shapes the investigation more than the model.} Comparing the behavior profiles of
runs (cosine similarity of their behavior histograms), two different models on the same finding are
twice as similar (0.58) as one model on two different findings (0.29); unrelated runs score 0.23.

\textbf{Models that disagree investigated the same way.} Within each finding, pairs of models that
reach opposite verdicts are less similar in behavior than pairs that agree by only 0.022 (95\%
interval 0.009--0.034), a fraction of the gap between task and model. On one container finding,
Claude and DeepSeek read the same three compose files in the same order and query the same pattern;
from the same evidence one calls an unrestricted database volume a vulnerability, the other a false
positive. With errors shared across models (Appendix~\ref{app:sastbench-sql}), disagreement and error
live in the judgment, not in the investigation.

\subsection{Anomalies as a spotlight}
\label{app:sastbench-anomalies}

Outliers in the discretized runs are not failures in themselves, but they point at subsets worth
reading, and some of those are fixable.

\begin{itemize}
  \item \textbf{Unusual runs go to unusual places.} The runs least likely under each agent's own
    transition model read test files, templates and configuration; most end in a correct
    ``non-production code'' dismissal.
  \item \textbf{Failed lookups are evidence, but not for every agent.} Rare transitions into a
    verdict lead to verdicts issued right after a failed read or search (4--14\% of verdicts).
    Claude, DeepSeek and Gemini 2.5 Pro then call ``vulnerable'' less (38\% against 47--50\%), and
    only on false-positive findings (47\% to 36\%; unchanged on true positives), using the empty result
    as evidence. Gemini 3 Pro does not adjust, even after its own malformed calls.
  \item \textbf{Runs that hit the step limit are symbol hunts.} They over-represent symbol-definition
    searches 2.1--2.9$\times$ and repeat a behavior longer; for Gemini 2.5 Pro a fifth of their steps
    are error recovery. The fix suggested is in the symbol-lookup tool, not the model.
\end{itemize}

\subsection{Building a new taxonomy with a decision model}
\label{app:sastbench-mistakes}

The construction of Appendix~\ref{app:jev-construction} builds taxonomies Taxi was not given. Here it
builds one of the agents' local mistakes. A decision model scored 11,760 sampled steps for a local
mistake, seeing each step with the earlier calls and their outputs, the tool's output and the agent's
next reasoning (paths, file names and languages masked); construction then built categories over
the 594 highest-scoring steps, with guidance never to use the vulnerability class, repository or
language. Its 28 categories group into four families (Table~\ref{tab:sastbench-mistakes}), for
example a failed search followed by ``I have completed the analysis and confirmed the
vulnerability'', or ``no security patterns found'' read as ``this confirms the absence of integrity
verification''. Scoring took 11,760 decision-model calls (under 5 minutes) and construction about 8
minutes.

\begin{table}[h]
\centering
\small
\begin{tabular}{@{}p{0.42\linewidth}rrrr@{}}
\toprule
Per 1,000 steps & Claude & Gemini 3 & Gemini 2.5 & DeepSeek \\
\midrule
Wrong tool arguments (nonexistent path or glob, no line range on a large file, minified files) & 8.7 & 19.3 & 7.0 & 12.4 \\
Repeating a call unchanged after it failed or answered & 2.6 & 11.0 & 11.4 & 10.5 \\
Concluding from a failed or empty lookup & 0.0 & 3.9 & 3.1 & 5.6 \\
Inefficient investigation (generic searches, reading boilerplate, reading before locating) & 16.8 & 13.6 & 17.6 & 27.0 \\
\bottomrule
\end{tabular}
\caption{Local mistakes per 1,000 steps, from the decision-model-built taxonomy. Lower bounds: only
the highest-scoring 5\% of steps was categorized.}
\label{tab:sastbench-mistakes}
\end{table}

\section{Related Work}
\label{app:related_work}
Agent trajectory clustering and its taxonomization have been studied extensively. For example, \cite{wan2024tnt,lam2024concept,tamkin2024clio}, and others cluster conversations and create hierarchical taxonomies with natural language descriptions. Their focus was on taxonomizing trajectories as a whole, while our work operates at the granularity of steps, closer to \cite{tyagi2026globalcot}, though they worked on a single input prompt at a time. That work belongs to the field of interpretability: it targeted chains-of-thought of simple problem solving tasks in order to understand the various methods a model takes in solving the problem. More broadly, LLMs have been used to induce topics and goal-driven clusters described in natural language \citep{pham2024topicgpt, wang2023goal}, to generate and validate taxonomies with humans in the loop \citep{shah2025using}, and to organize a given set of entities into a taxonomy \citep{zeng2024chain}. A related line of work uses LLMs to describe in natural language how two text corpora differ \citep{zhong2022describing, zhong2023goal}, and to fit statistical models, including clustering, whose parameters are natural-language predicates rather than embedding centroids \citep{zhong2024explaining}.

TypeSafe AI introduced Jev as a ``System One'' model for decisions inside software
\citep{almeida2026systemone}. Rather than produce free-form text, it accepts application state and
typed questions and returns choices, scores, or yes/no probabilities. The interface is intended for
classification, routing, filtering, and gating at substantially lower cost than generative-model
calls. Taxi adopts the interface rather than depending on the original model: throughout this report,
Jev-style includes any compatible decision model capable of answering these bounded questions.

Over the years, many agent debugging tools have emerged. Laminar \citep{laminar2026} and Docent \citep{meng2025docent} are open source frameworks that use LLMs to identify problems in agentic traces. Recently, LangSmith announced its own LangSmith Engine \citep{tannyhill2026langsmith}, similarly addressing this problem. All of them now include querying abilities to varying degrees of expressivity. None of them uses clustering as a central step; they use it post hoc, to aggregate \textit{results only} for better human digestibility.

Another related line of work focuses on agent taxonomies \citep{cemri2026multi, gao2026interpret, ashury2026errormap}. Most of them are constructed manually or at least semi-manually. They are designed to capture behavior on a global level, not tailored to the specific dataset explored. 
Cemri et al \citep{cemri2026multi} focus on multi-agent failures. Gao et al \citep{gao2026interpret} focus on agent actions in general, closer in spirit to us, rather than just failures. Ashury-Tahan et al \citep{ashury2026errormap} focus on failures, and they propose ErrorAtlas, a closed taxonomy, and ErrorMap, a taxonomy construction pipeline based on LLMs. 
Closer to Taxi in that the taxonomy is tailored to the data, AdaMAST \citep{cemri2026adaptive} uses LLMs to induce a failure taxonomy from an agent system's own execution traces, and Gupta et al \citep{gupta2025unsupervised} discover failure taxonomies from robot deployment logs by having a vision-language model explain each failed trajectory and clustering the explanations. Both target failures rather than behavior in general, and both rely on a model reading the trajectories, whereas Taxi's embedding and regex paths assign steps to clusters without any model call.

It will be interesting in the future to absorb such a taxonomy into Taxi, as a set of \textit{pre-defined constructs}, that Taxi will then modify or refine for the specific use case.

\section{Advanced: Database Schema}
\label{app:schema}

Every table below lives in the single database described in Section~\ref{sec:backbone};
\texttt{pk}/\texttt{fk} mark primary/foreign keys, and \texttt{jsonb}/\texttt{vector} mark
JSON and vector columns respectively. Foreign keys not marked otherwise cascade on
delete.

\paragraph{Corpus (data-source scope).}
\begin{description}
\item[\texttt{data\_sources}] \texttt{id} (uuid, pk), \texttt{name}, \texttt{source\_type},
  \texttt{source\_spec} (jsonb), \texttt{n\_trajectories}, \texttt{n\_steps},
  \texttt{data\_version}, \texttt{eval\_data\_version}, \texttt{embed\_model},
  \texttt{embed\_state} (jsonb), \texttt{created\_at}, \texttt{updated\_at}. One row per ingested
  corpus; \texttt{data\_version}/\texttt{eval\_data\_version} are the watermarks every incremental
  consumer (embedding, evaluators) reads against.
\item[\texttt{trajectories}] \texttt{id} (text, pk), \texttt{data\_source\_id} (fk
  $\to$ \texttt{data\_sources}), \texttt{n\_steps}, \texttt{meta} (jsonb), \texttt{source\_key},
  \texttt{created\_at}. One row per agent run; \texttt{meta} carries source-specific fields (e.g.\
  an external run id, the ingest data version it arrived under); \texttt{source\_key} is the
  stable external identifier incremental re-sync deduplicates against.
\item[\texttt{steps}] \texttt{id} (bigint, pk), \texttt{trajectory\_id} (fk
  $\to$ \texttt{trajectories}), \texttt{step\_index}, \texttt{role}, \texttt{action},
  \texttt{observation}, \texttt{tool\_calls} (jsonb), \texttt{system\_prompt},
  \texttt{message\_extras} (jsonb), \texttt{summary\_text}, \texttt{embedding} (vector). One row
  per message/tool-call/observation within a trajectory; \texttt{embedding} is what
  embedding-based clustering and semantic search consume.
\end{description}

\paragraph{Taxonomy (project scope).}
\begin{description}
\item[\texttt{projects}] \texttt{id} (uuid, pk), \texttt{name}, \texttt{owner},
  \texttt{data\_source\_id} (fk $\to$ \texttt{data\_sources}, restrict), \texttt{created\_at},
  \texttt{taxonomy\_settings} (jsonb), \texttt{pinned\_hints} (jsonb), \texttt{filters} (jsonb),
  \texttt{synced\_data\_version}, plus a cached-mean-length validity pair. One row per analysis
  project; \texttt{filters} defines the project's view-level scope (Section~\ref{sec:backbone});
  \texttt{synced\_data\_version} is the data version that was last synced to the project.
\item[\texttt{project\_collaborators}] \texttt{project\_id} (pk, fk $\to$ \texttt{projects}),
  \texttt{email} (pk), \texttt{created\_at}. Per-project access list.
\item[\texttt{taxonomy\_versions}] \texttt{id} (int, pk), \texttt{project\_id} (fk
  $\to$ \texttt{projects}), \texttt{version}, \texttt{created\_at}, \texttt{k},
  \texttt{pipeline\_id}, \texttt{settings} (jsonb). One immutable row per published taxonomy
  snapshot.
\item[\texttt{clusters}] \texttt{id} (int, pk), \texttt{taxonomy\_version\_id} (fk
  $\to$ \texttt{taxonomy\_versions}), \texttt{cluster\_id}, \texttt{label}, \texttt{description},
  \texttt{kind} (regex, embedding, or jev), \texttt{n\_members}, \texttt{mean\_chars},
  \texttt{centroid} (vector), \texttt{pattern} (text,
  nullable regex), \texttt{parent\_cluster\_id} (fk $\to$ \texttt{clusters}, nullable),
  \texttt{lft}, \texttt{rgt}. One row per cluster in a taxonomy version's tree;
  \texttt{lft}/\texttt{rgt} are a nested-set encoding, so ``member of cluster $C$'' is a single
  range check rather than a recursive query.
\item[\texttt{step\_assignments}] \texttt{taxonomy\_version\_id} (pk, fk), \texttt{trajectory\_id}
  (pk, fk), \texttt{step\_index} (pk), \texttt{cluster\_id}. Composite-keyed step$\to$cluster
  mapping for one taxonomy version.
\item[\texttt{taxonomy\_pipeline\_state}] \texttt{project\_id} (fk), \texttt{pipeline\_id},
  \texttt{labels} (int array), \texttt{cluster\_labels} (jsonb), \texttt{silhouette\_state}
  (jsonb), \texttt{centroids} (jsonb), \texttt{step\_refs} (jsonb), \texttt{created\_at}.
  Transient, resumable staging state for an in-flight construction pipeline -- the $k$-selection
  search checkpoint and incremental per-cluster labeling progress.
\end{description}

\paragraph{Chat.}
\begin{description}
\item[\texttt{conversations}] \texttt{id} (uuid, pk), \texttt{project\_id} (fk
  $\to$ \texttt{projects}), \texttt{title}, \texttt{created\_at}, \texttt{updated\_at}.
\item[\texttt{messages}] \texttt{id} (uuid, pk), \texttt{conversation\_id} (fk
  $\to$ \texttt{conversations}), \texttt{role}, \texttt{content}, \texttt{created\_at},
  \texttt{metadata} (jsonb).
\end{description}
Chat history backing the interactive corpus-debugging surfaces.

\paragraph{Job engine.}
\begin{description}
\item[\texttt{jobs}] \texttt{id} (uuid, pk), \texttt{data\_source\_id} (fk, nullable),
  \texttt{project\_id} (fk, nullable), \texttt{kind}, \texttt{status}, \texttt{phase},
  \texttt{meta} (jsonb), \texttt{paused}, \texttt{pipeline\_id}, \texttt{pipeline\_superseded},
  \texttt{retry\_count}, \texttt{started\_at}, \texttt{updated\_at}, \texttt{finished\_at}. One
  row per long-running operation.
\item[\texttt{job\_events}] \texttt{id} (bigint, pk), \texttt{job\_id} (fk $\to$ \texttt{jobs}),
  \texttt{seq}, \texttt{kind}, \texttt{message}, \texttt{payload} (jsonb), \texttt{ts}. The
  append-only, ordered event log per job (phase/progress/thinking/tool/error events) backing
  streaming progress and after-the-fact inspection.
\end{description}

\paragraph{Evaluators and insights.}
\begin{description}
\item[\texttt{evaluators}] \texttt{id} (text, pk), \texttt{yaml}, \texttt{enabled},
  \texttt{created\_at}, \texttt{updated\_at}. One row per published evaluator; \texttt{id} is the
  evaluator name and \texttt{yaml} is the complete definition described in
  Section~\ref{sec:ecosystem}.
\item[\texttt{insight\_definitions}] \texttt{id} (text, pk), \texttt{yaml}, \texttt{enabled},
  \texttt{created\_at}, \texttt{updated\_at}. Same shape as \texttt{evaluators}, for insight
  definitions.
\item[\texttt{insight\_reports}] \texttt{id} (uuid, pk), \texttt{insight\_id} (fk
  $\to$ \texttt{insight\_definitions}), \texttt{data\_source\_id} (fk), \texttt{data\_version},
  \texttt{trajectory\_count}, \texttt{report\_markdown}, \texttt{summary\_markdown},
  \texttt{slack\_message\_ts}, \texttt{slack\_thread\_posted}, \texttt{created\_at}.
\item[\texttt{insight\_reports\_dry\_run}] a schema-identical shadow of
  \texttt{insight\_reports} that dry-run report generation writes to instead, for tests that don't touch production reports or their watermark.
\end{description}

\paragraph{Shared, scoped read views.}
Rather than querying the tables above directly, every consumer is encouraged to read through four views,
re-created at startup and parameterized per request by session-local settings (which
project or data source(s) are in scope, the project's own filters, and, for an insight run, its
analysis-window bounds):
\texttt{project\_trajectories(id, n\_steps, meta, created\_at, data\_source\_id)},
\texttt{project\_steps(id, trajectory\_id, step\_index, role, action, observation, tool\_calls,
system\_prompt, summary\_text, data\_source\_id)},
\texttt{project\_clusters(cluster\_id, label, description, kind, pattern, n\_members, mean\_chars, lft,
rgt, parent\_cluster\_id)}, and
\texttt{project\_step\_assignments(trajectory\_id, step\_index, cluster\_id)}. This is an instance
of the shared-backbone design in Section~\ref{sec:backbone}: scoping logic
is written once, rather than once per consumer.

\end{document}